\documentclass{salesforce_ai_research}
\usepackage{amsmath}
\usepackage{amssymb}
\usepackage{graphicx}
\usepackage{booktabs}
\usepackage{lipsum}
\usepackage{newtxmath}
\newcommand{\eg}{\emph{e.g.}}
\newcommand{\ie}{i.\,e.}
\renewcommand{\paragraph}[1]{\noindent\textbf{\color{salesforcedarkblue}{#1}}}
\usepackage{booktabs}
\usepackage{tabularx}
\newcolumntype{Y}{>{\centering\arraybackslash}X}
\usepackage{multicol}
\usepackage{multirow}
\usepackage{xcolor}
\usepackage{colortbl}
\crefname{figure}{Fig.}{Figs.}
\Crefname{figure}{Figure}{Figures}
\crefname{section}{Sec.}{Secs.}
\Crefname{section}{Section}{Sections}
\Crefname{table}{Table}{Tables}
\crefname{table}{Tab.}{Tabs.}
\Crefname{equation}{Equation}{Equations}
\crefname{equation}{Eq.}{Eqs.}
\Crefname{appendix}{Appendix}{Appendixs}
\crefname{appendix}{App.}{Apps.}
\definecolor{lightgreen}{RGB}{179,216,130}
\usepackage[position=bottom]{subfig}
\usepackage{pgfplots}
\pgfplotsset{compat = newest}
\usepgfplotslibrary{dateplot}   
\usepgfplotslibrary{groupplots}
\usepgfplotslibrary{groupplots}
\usepackage{fancyvrb} 
\usepackage{cuted} 
\usepackage{bbding}
\usepackage{makecell}
\usepackage{xspace}
\NewDocumentCommand\DeclareImageMark{m m O{6pt}}{
  \pgfdeclareimage[height=#3,width=#3]{#1-image}{#2}
  \expandafter\pgfdeclareplotmark\expandafter{#1}{%
    \pgfpathcircle{\pgfpointorigin}{0pt}
    \pgftext[at=\pgfpointorigin,center]{\pgfuseimage{#1-image}}
  }
}
\DeclareImageMark{uitars}{fig/logo/uitars}[6.2pt] 
\DeclareImageMark{qwen}{fig/logo/qwen}[6.2pt] 
\DeclareImageMark{uground}{fig/logo/uground}[6.2pt] 
\DeclareImageMark{infigui}{fig/logo/infigui}[6.2pt]
\DeclareImageMark{segui}{fig/logo/segui}[6.2pt]
\DeclareImageMark{jedi}{fig/logo/jedi}[6.2pt]
\DeclareImageMark{openai}{fig/logo/openai}[6.2pt]
\DeclareImageMark{claude}{fig/logo/claude}[6.2pt]
\DeclareImageMark{gemini}{fig/logo/gemini}[6.2pt]
\DeclareImageMark{atlas}{fig/logo/atlas}[6.2pt]
\usepackage{xcolor}

\newcommand{\name}{GTA1\xspace}
\usepackage{colortbl}
\usepackage{float}

\usepackage[subfigure]{tocloft}
\usepackage{etoc}
\usepackage{titletoc}
\newcommand\DoToC{
  \startcontents
  \printcontents{}{1}{\noindent \textbf{\Large{Table of Contents in Appendix}}\vskip3pt\vskip5pt}
  \vskip3pt\vskip5pt
}

\usepackage[most]{tcolorbox}
\tcbuselibrary{skins}
\tcbuselibrary{breakable}
\usepackage[smartEllipses]{markdown}
\usepackage{listings}
\usepackage{listingsutf8}
\usepackage{longtable}
\UseRawInputEncoding
\usepackage[T1]{fontenc}
\renewcommand{\cite}{\citep}
\title{GTA1: GUI Test-time Scaling Agent}

\author{
\centering
\textbf{Yan Yang}$^{1}$\and
\textbf{Dongxu Li}~\Envelope$^{2}$\and 
\textbf{Yutong Dai}$^{1}$\and 
\textbf{Yuhao Yang}$^{3}$\and 
\textbf{Ziyang Luo}$^{1}$\and 
\textbf{Zirui Zhao}$^{1}$\and
\textbf{Zhiyuan Hu}$^{1}$\and
\textbf{Junzhe Huang}$^{2}$\and
\textbf{Amrita Saha}$^{1}$\and
\textbf{Zeyuan Chen}$^{1}$\and
\textbf{Ran Xu}$^{1}$\and
\textbf{Liyuan Pan}$^{2}$\and
\makebox[\textwidth][c]{\textbf{Silvio Savarese} \quad \textbf{Caiming Xiong}~\Envelope$^{1}$ \quad \textbf{Junnan Li}~\Envelope$^{1}$}\\
\makebox[\textwidth][c]{
$^1$Salesforce AI Research \quad $^2$The Australian National University}\\
\makebox[\textwidth][c]{$^3$University of Hong Kong \quad \Envelope~Corresponding Author}\\
{\tt\small dongxuli1005@gmail.com \quad cxiong@salesforce.com \quad junnan.li@salesforce.com}
}

\begin{document}
\arrayrulecolor{salesforcedarkblue}
\maketitle

\begin{abstract}
Graphical user interface (GUI) agents autonomously complete tasks across platforms (\eg, Linux) by sequentially decomposing user instructions into action proposals that iteratively interact with visual elements in the evolving environment. However, two main challenges arise:
i) planning (\ie, the action proposal sequence) under expansive action space, where selecting an appropriate plan is non-trivial, as many valid ones may exist; 
ii) accurately grounding actions in complex and high-resolution interfaces, \ie, precisely interacting with visual targets.
This paper investigates the aforementioned challenges with our \textbf{G}UI \textbf{T}est-time Scaling \textbf{A}gent, namely \name. First, we conduct test-time scaling to select the most appropriate action proposal: at each step, multiple candidate proposals are sampled and evaluated and selected by a judge model. It trades off computation for better decision quality by concurrent sampling.
Second, we propose a model that improves grounding of the selected action proposals to its corresponding visual elements. Our key insight is that reinforcement learning (RL) facilitates grounding through inherent objective alignments, rewarding successful clicks on interface elements.
Experimentally, \name achieves state-of-the-art performance on both grounding and agent task execution benchmarks. The code and models are released \href{https://github.com/Yan98/GTA1}{here}. 
\end{abstract}

\section{Introduction}

Automating task completions across diverse platforms through GUI agents represents a significant milestone toward general artificial intelligence, supporting activities from online orders to expert workflows \cite{yang2024aria}. To solve a task, a GUI agent translates user instructions into multi-step interactions such as action proposals consisting of clicks or keystrokes \cite{gou2024navigating}. This introduces a planning challenge, as multiple valid action proposal sequences may exist for the same user task. 
The challenge is further amplified by the high-resolution (up to 4K), complex, and hierarchical layouts of GUI \cite{li2025screenspotpro,wu2024atlas,cheng2024seeclick,xie2025scaling},  requiring accurate coordinate identification of the target interface elements. This work aims to address both challenges (\ie, planning and grounding) towards a performant GUI agent.

Formally, existing works \cite{yang2024aria,agashe2024agent,agashe2025agent,xie2025scaling,xu2024aguvis} often pair a GUI grounding model with a planner (\eg, o3 \cite{openAI_o3_o4_mini}).
The planner determines an action proposal at each step, while the grounding model locates the target interface elements for interactions (\eg, click areas). However, due to the inherent flexibility of user tasks, multiple feasible action proposal sequences may exist for completing the same task, some more direct and efficient than others. This makes the agent vulnerable to cascading failures, \ie, errors in early grounding or planning steps can derail the entire task. One way to avoid this is to roll out full action sequences in advance, but unlike domains such as math problem-solving, GUI environments lack a ``lookahead'' capability: actions often have irreversible state effects, limiting the practicality in real-world use.   This raises a central question: \textit{how can GUI agents remain robust in planning despite the lack of ``lookahead'' and the presence of multiple plausible action proposal sequences?} 

\begin{figure}[!t]
\includegraphics{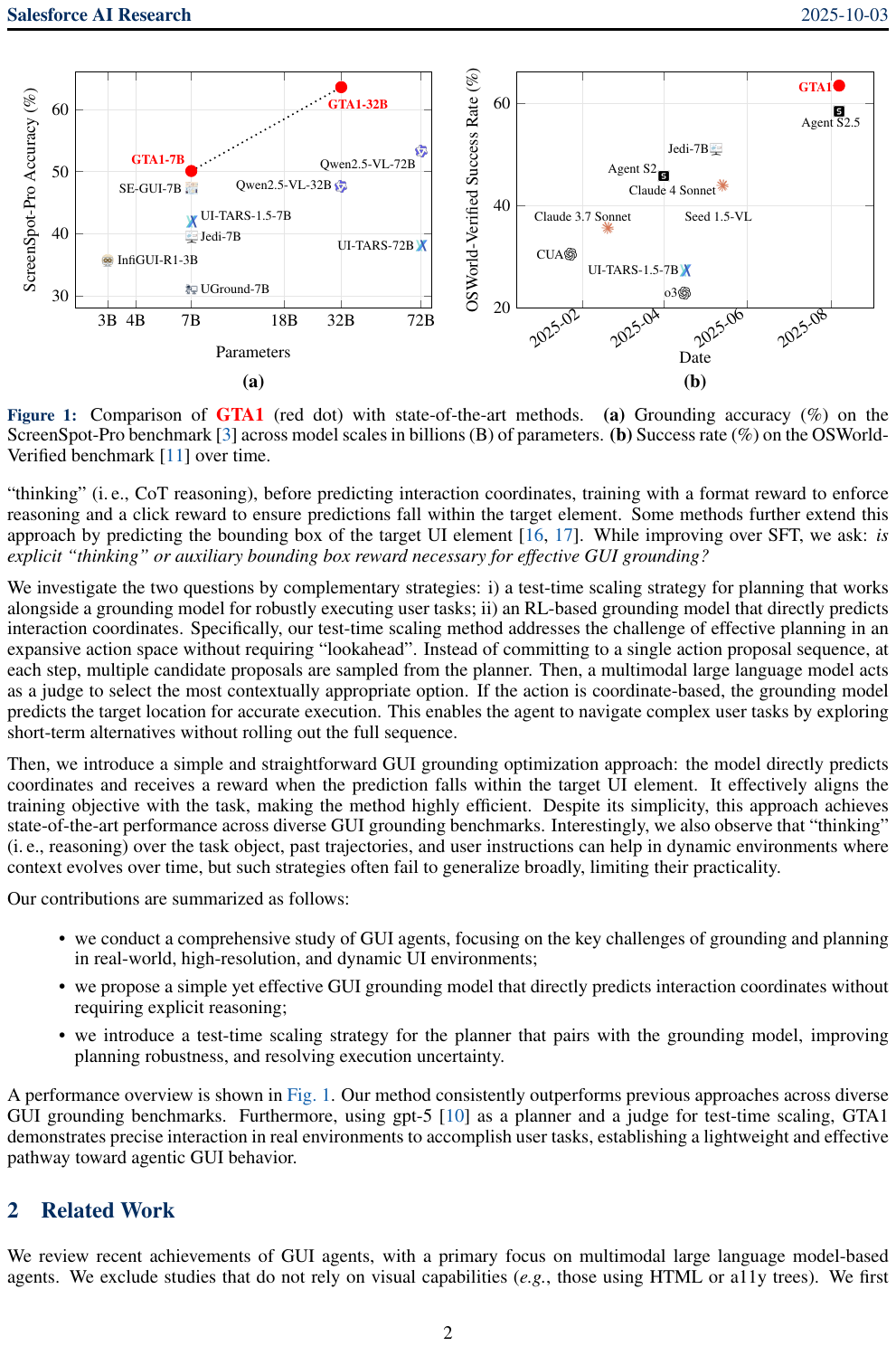}
  \vspace{-.5em}
  \caption{Comparison of \textbf{\textcolor{red}{GTA1}} (red dot) with state-of-the-art methods. \textbf{(a)} Grounding accuracy (\%) on the ScreenSpot-Pro benchmark \cite{li2025screenspotpro} across model scales in billions (B) of parameters. \textbf{(b)} Success rate (\%) on the OSWorld-Verified benchmark \cite{xie2024osworld} over time.}
  \label{fig:overview}
\end{figure}

Beyond planning, GUI grounding models predominantly rely on supervised fine-tuning (SFT) \cite{yang2024aria, cheng2024seeclick, gou2024navigating, wu2024atlas, xie2025scaling}, which trains models to predict the exact center of target interface elements. While effective in simple settings, this approach struggles to generalize to complex and high-resolution scenarios, particularly in professional interfaces \cite{li2025screenspotpro}. Fundamentally, SFT introduces a misalignment with the task itself: \textit{any coordinate within the target element constitutes a valid interaction}, yet SFT penalizes deviations from the center. It limits model flexibility, reduces robustness, and preventing the model from perceiving  appropriate supervision signals.

Alternatively, inspired by DeepSeek-R1-Zero \cite{guo2025deepseek}, RL, particularly Group Relative Policy Optimization (GRPO) \cite{shao2024deepseekmath}, has been explored in GUI grounding. Following \cite{guo2025deepseek},  prior works \cite{luo2025gui, lu2025ui, liu2025infigui} formulates by performing a textual ``thinking'' (\ie, CoT reasoning), before predicting interaction coordinates, training with a format reward to enforce reasoning and a click reward to ensure predictions fall within the target element. Some methods further extend this approach by predicting the bounding box of the target UI element \cite{liu2025infigui, zhou2025gui}. While improving over SFT, we ask: \textit{is explicit ``thinking'' or auxiliary bounding box reward  necessary for effective GUI grounding?}

We investigate the two questions by complementary strategies:
i) a test-time scaling strategy for planning that works alongside a grounding model for robustly executing user tasks;
ii) an RL-based grounding model that directly predicts interaction coordinates. 
Specifically, our test-time scaling method addresses the challenge of effective planning in an expansive action space without requiring ``lookahead''. Instead of committing to a single action proposal sequence, at each step, multiple candidate proposals are sampled from the planner. Then, a multimodal large language model acts as a judge to select the most contextually appropriate option. If the action is coordinate-based, the grounding model predicts the target location for accurate execution. This enables the agent to navigate complex user tasks by exploring short-term alternatives without rolling out the full sequence.

Then, we introduce a simple and straightforward GUI grounding optimization approach: the model directly predicts coordinates and receives a reward when the prediction falls within the target UI element. It effectively aligns the training objective with the task, making the method highly efficient. Despite its simplicity, this approach achieves state-of-the-art performance across diverse GUI grounding benchmarks. Interestingly, we also observe that ``thinking'' (\ie, reasoning) over the task object, past trajectories, and user instructions can help in dynamic environments where context evolves over time, but such strategies often fail to generalize broadly, limiting their practicality.

Our contributions are summarized as follows:
\begin{itemize}
    \item we conduct a comprehensive study of GUI agents, focusing on the key challenges of grounding and planning in real-world, high-resolution, and dynamic UI environments;
    
    \item we propose a simple yet effective GUI grounding model that directly predicts interaction coordinates without requiring explicit reasoning;
    
    \item we introduce a test-time scaling strategy for the planner that pairs with the grounding model, improving planning robustness, and resolving execution uncertainty.
\end{itemize}
A performance overview is shown in \cref{fig:overview}. Our method consistently outperforms previous approaches across diverse GUI grounding benchmarks. Furthermore, using gpt-5 \cite{openAI_o3_o4_mini} as a planner and a judge for test-time scaling, \name demonstrates precise interaction in real environments to accomplish user tasks, establishing a lightweight and effective pathway toward agentic GUI behavior.

\section{Related Work}
We review recent achievements of GUI agents, with a primary focus on multimodal large language model-based agents. We exclude studies that do not rely on visual capabilities (\eg, those using HTML or a11y trees).  We first reviews the GUI grounding, and then categorize existing agents into two broad types: native GUI agents and two-stage GUI agents.

\paragraph{GUI Grounding.} GUI grounding refers to the task of mapping user instructions to specific coordinates corresponding to target UI elements. Early works \cite{yang2024aria, cheng2024seeclick, xu2024aguvis, gou2024navigating} primarily focus on SFT, training models to predict the center point of the intended UI element. However, SFT does not fully align with the nature of the GUI grounding task, where any coordinate within the target element should be considered a success. As a result, SFT-based models often exhibit poor generalization (\eg, on high-resolution and visually complex user interfaces \cite{li2025screenspotpro}). With the success of DeepSeek-R1-Zero \cite{guo2025deepseek}, RL (specifically GRPO \cite{shao2024deepseekmath}) has drawn increased attention. Many recent efforts naively replicate techniques from other domains \cite{luo2025gui,lu2025ui,liu2025infigui}, such as prompting the model to generate a ``thinking'' (\ie, CoT reasoning) before producing an answer, and the answer is rewarded only if the predicted coordinates fall within the target element region.  This strategy overlooks an important insight: the ``thinking'' degrades performance in GUI grounding.
A concurrent study \cite{zhou2025gui} makes similar observations, noting that CoT reasoning (\ie, ``thinking'') is not required for RL training in GUI grounding and may even hinder grounding accuracy. Our work further distinguishes itself in the following ways:
i) we clarify that ``thinking'' is not necessary for GUI grounding in static environments;
ii) we demonstrate that ``thinking'' improves grounding performance in dynamic, real-world environments when provided with past trajectories and task objectives;
iii) we conduct a comprehensive study of RL-based GUI grounding across models of various scales. Going beyond prior work, we further evaluate how the model, when paired with a planner, performs in realistic and dynamic environments, a critical aspect that existing studies largely overlook.

\paragraph{Two-stage GUI Agent.} One major challenge in GUI grounding is accurately locating the coordinates of UI elements intended for interaction. To address this, two-stage GUI agents modularize into planning and action, each handled by separate models \cite{gou2024navigating}. They usually leverage advanced reasoning models, such as GPT-4o \cite{hurst2024gpt} and Claude 3.7 \cite{claude37}, as planners to generate an action proposal for each step from the user task instruction, using real-time UI screenshot and past trajectories as context. A separate grounding module then maps these instructions to specific UI elements, enabling the development of vision-only GUI agents \cite{cheng2024seeclick,yang2024aria,gou2024navigating}. While most existing work primarily focuses on GUI grounding, more complex components, such as memory management and external knowledge bases, are also being explored to enhance agent performance \cite{agashe2025agent}. This paper follows the two-stage method on establishing a GUI agent.

\paragraph{Native GUI Agent.} A native GUI agent completes user tasks in an end-to-end manner. Four main aspects are studied \cite{qin2025ui}:
i) perception,  interpreting the user interface to understand the current state;
ii) memory, storing knowledge and historical experiences to support  making decisions;
ii) planning, analyzing the task and reflecting on progress to generate action proposals;
iv) action, performing atomic operations based on the action proposal to effectively progress toward the task goal. Examples of native GUI agents can be specified to CUA \cite{hurst2024gpt,openAI_o3_o4_mini} and Claude Computer Use \cite{claude37}. One of the main challenges for native GUI agents lies in long-context learning. To address this, some approaches employ a sliding window mechanism \cite{qin2025ui}, while others maintain a textual description of past trajectories to manage contextual information \cite{xu2024aguvis}. In practice, end-to-end native GUI agents have demonstrated strong performance in completing user tasks, as shown by benchmarks that reflect dynamic and realistic UI environments, such as OSWorld \cite{xie2024osworld}.  However, this paper is the first to show that a two-stage GUI agent can achieve competitive performance in such environments, challenging the assumption that end-to-end approaches are inherently superior.

\section{Method}

\paragraph{Overview.} Our method adopts a two-stage GUI agent framework composed of a planner and a grounding model, and focuses on improving planning robustness and grounding accuracy through the following key components: i) test-time scaling for planning, which scales inference computation to effectively handle planning selection challenges in complex GUI environments;
ii) grounding model training, filtering out training samples with annotation errors to improve supervision quality, and optimizing a grounding model using RL (\eg, GRPO) to directly predict coordinates without relying on any intermediate ``thinking'' (\ie, CoT reasoning) on the derived data..

\begin{figure}[!t]
    \centering
    \includegraphics[width=\linewidth]{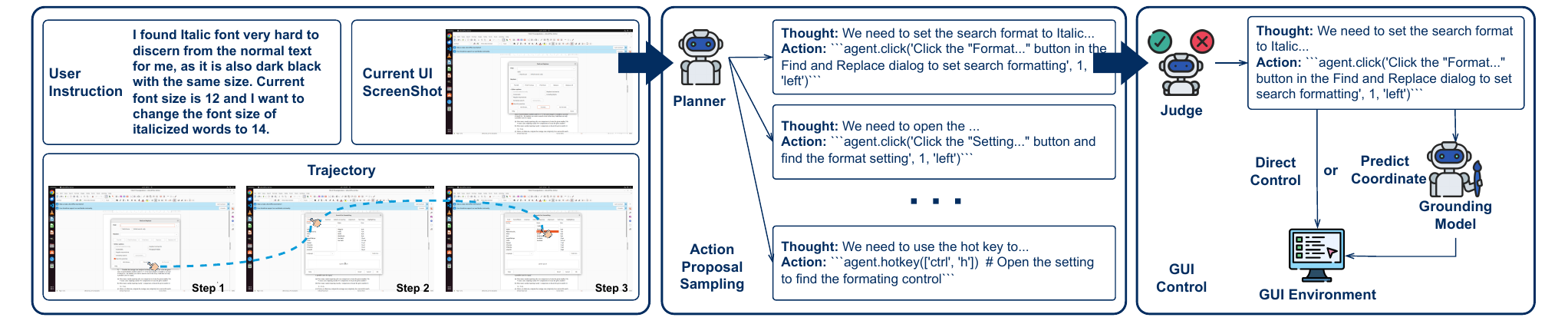}
    \vspace{-1em}
    \caption{Overview of our GUI agent architecture. At each step, the trajectory, current UI screenshot, and user instruction are sent to a planner, which samples multiple action proposals. A multimodal large language model judge is then used to select the best candidate action proposal.  When the candidate action proposal is a coordinate-based action (\eg, a click), the grounding model predicts a precise interaction point on the GUI for executing the action. For non-coordinate-based actions (\eg, key presses), the action can be executed directly without grounding.
    }
    \label{fig:gui_agent}
\end{figure}

\subsection{Test-time Scaling for Planning}

At each step of executing user instructions in real-world environments, a planner is provided with the user instruction (\ie, task objective), the trajectory so far, and the current UI screenshot. Based on this context, we sample $K$ candidate action proposals, denoted as $\{\boldsymbol{p}_{k}\}_{k=1}^{K}$, where  $\boldsymbol{p}_{k}$ represents a corresponding action (\eg, clicking `the blue button' or performing a keystroke).

Then, a multimodal large language model judge is used to evaluate the $K$ candidates $\{\boldsymbol{p}_{k}\}_{k=1}^{K}$ based on their alignment with the user intent and the current GUI state. The judge, which can be the planner model itself, picks the best candidate action proposal $\boldsymbol{p}_{k*}$ from $\{\boldsymbol{p}_{k}\}_{k=1}^{K}$, allowing the agent to select the most contextually appropriate option. Once the best candidate action proposal $\boldsymbol{p}_{k*}$ is selected, the grounding model $\pi(\cdot, \cdot)$ takes $\boldsymbol{p}_{k*}$ and the current screenshot $\boldsymbol{s}$ as the input. If $\boldsymbol{p}_{k*}$ is a coordinate-based action (\eg, a click), the grounding model predicts a precise interaction point on the GUI, which is then used to execute the action. For non-coordinate-based actions (\eg, key presses or text input), the action can be executed directly without grounding. This process is repeated step by step until the task is completed or the agent reaches a termination condition.

By incorporating sampling, judging, and grounding into each step, our agent avoids overcommitting to suboptimal action proposals and demonstrates improved robustness in complex and dynamic GUI environments. We show an overview in \cref{fig:gui_agent}.

\subsection{Grounding}
\paragraph{Data Cleaning.} We leverage curated open-source datasets to train our model, \eg, the Aria-UI collection \cite{yang2024aria}.
To train with a reward signal that verifies whether the predicted coordinates fall within the target UI element, we require a dataset that provides accurate bounding boxes for annotated interactive elements. For data points from desktop and web domains, the collection usually involves screenshots paired with accessibility tools such as A11y or HTML parsers to extract element structure and bounding box annotations. However, these bounding boxes can sometimes be misaligned with the actual visual rendering due to UI animations, timing inconsistencies, or rendering delays, introducing noise into the training signal. 

Therefore, to improve data quality, we apply a lightweight cleaning strategy. Given a UI screenshot $\boldsymbol{s}$, we use OmniParser \cite{lu2024omniparser}, denoted as $\mathcal{M}(\cdot)$, to detect all UI elements in $\boldsymbol{s}$, resulting in a set of bounding boxes $\{b_i\} = \mathcal{M}(\boldsymbol{s})$. Each $b_i$ represents the bounding box of a detected UI element.

\begin{figure}[!t]
    \centering
    \includegraphics[width=\linewidth]{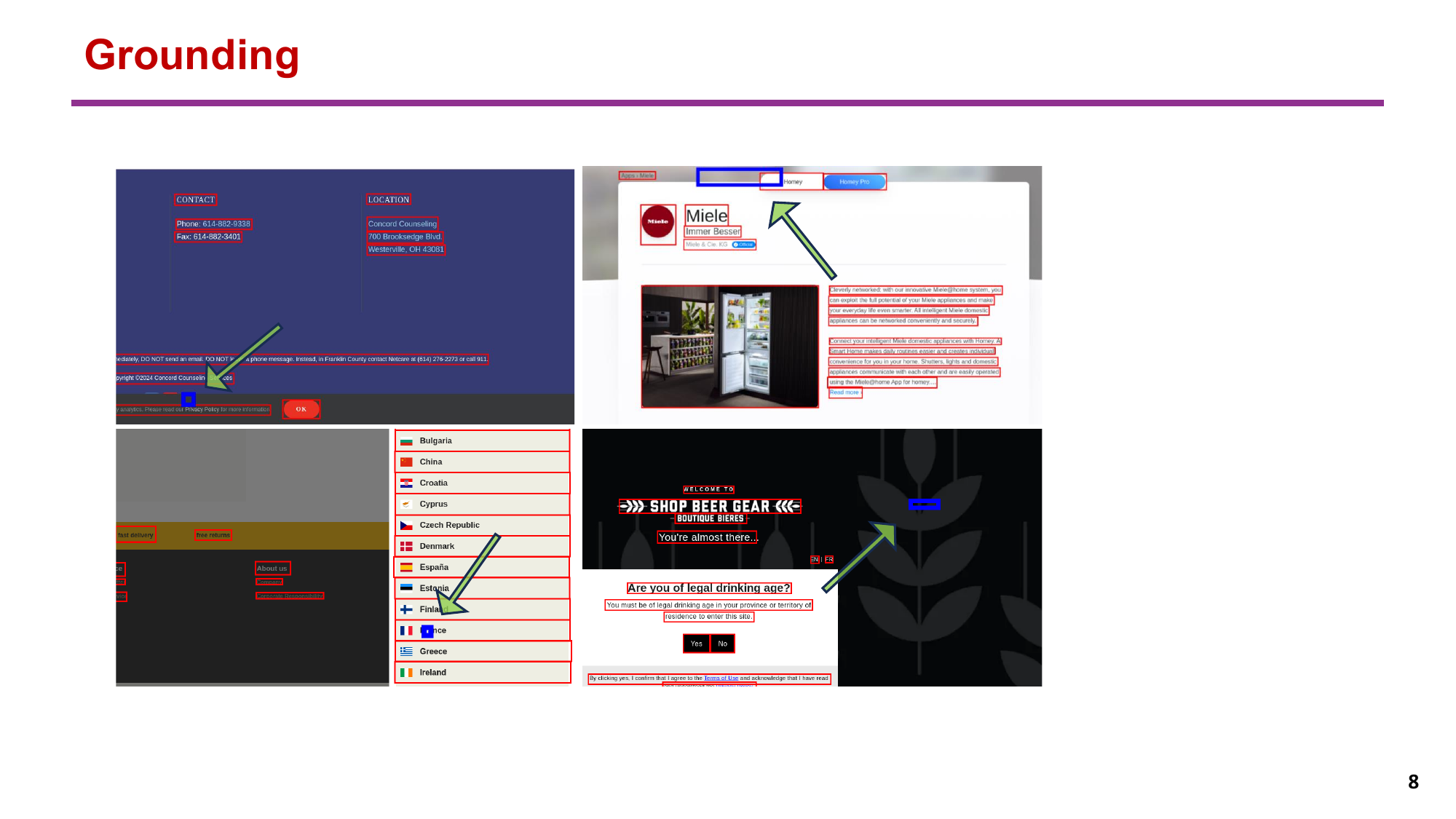}
    \caption{Examples from the Aria-UI dataset \cite{yang2024aria}. The \textcolor{blue}{blue bounding box} shows the annotation $b^{\mathsf{ann}}$, while \textcolor{red}{red bounding boxes} are detected by OmniParser~\cite{lu2024omniparser}. The \textcolor{lightgreen}{green arrow} highlights misaligned annotations, which our cleaning strategy filters out.}
    \label{fig:bbox}
\end{figure}

For a data point $\boldsymbol{s}$ with an annotated bounding box $b^{\mathsf{ann}}$, we discard the sample if the maximum Intersection over Union between $b^{\mathsf{ann}}$ and any $b_i \in \{b_i\}$ is smaller than a predefined threshold $\tau$,
\begin{align}
\max_{b_i \in \mathcal{M}(\boldsymbol{s})} \text{IoU}(b^{\mathsf{ann}}, b_i) < \tau \ ,
\end{align}
where $\text{IoU}(\cdot, \cdot)$ computes the overlap between two bounding boxes, defined as the area of their intersection divided by the area of their union. This helps ensure that the annotation $b^{\mathsf{ann}}$ in the training data remains consistent with the actual visual target, thereby reducing noise caused by misaligned annotations. We show some samples in \cref{fig:bbox}.

\paragraph{Training.}~In our RL training, we follow the GRPO framework \cite{guo2025deepseek, shao2024deepseekmath} to sample $N$ responses $\{\boldsymbol{o}_n\}_{n=1}^{N}$ from the policy multimodal large language model $\pi(\cdot, \cdot)$, given a screenshot $\boldsymbol{s}$ and an action proposal $\boldsymbol{p}$ as input. Here, each response $o_n$ represents a pair of pixel coordinates on the screen, \ie, $\boldsymbol{o}_n = (x_n,y_n)$, where $x_n$ and $y_n$ denote the horizontal and vertical positions, respectively.
Unlike prior approaches, we do not prompt the model to generate a ``thinking'' (\ie, CoT reasoning) before producing a response. Instead, the model directly outputs the predicted coordinates, aligning more closely with the nature of the GUI grounding task.

Then, each response is evaluated by checking whether the predicted coordinate $(x_n, y_n)$ falls within the annotated bounding box $\mathbf{b}^\mathsf{ann} = (x_{\text{min}}, y_{\text{min}}, x_{\text{max}}, y_{\text{max}})$. This yields a set of $N$ binary rewards $\{r_n\}_{n=1}^{N}$, where each reward is 
\begin{align}
r_n =
\begin{cases}
1, & \text{if } x_{\text{min}} \leq x_n \leq x_{\text{max}} \text{ and } y_{\text{min}} \leq y_n \leq y_{\text{max}} \ , \label{eq:reward} \\
0, & \text{otherwise} \ .
\end{cases}
\end{align}
We then normalize the rewards $\{r_n\}_{n=1}^{N}$ into advantages $\{A_n\}_{n=1}^{N}$ using Z-score normalization, 
\begin{align}
A_n = \frac{r_n - \frac{1}{N} \sum_{n=1}^{N} r_n}{\sqrt{\frac{1}{N} \sum_{n=1}^{N} (r_n - \frac{1}{N} \sum_{n=1}^{N} r_n)^2}} \ . 
\end{align} 
Finally, the model is optimized by 
\begin{align}
    \mathcal{L} = - \frac{1}{N} \sum_{n=1}^{N}  \min \Bigl(\frac{\pi(o_{n} \mid \boldsymbol{s}, \boldsymbol{p})}{\pi^\mathsf{old}(o_{n} \mid \boldsymbol{s}, \boldsymbol{p})} \cdot A_{n}, \text{clip}\bigl(\frac{\pi(o_{n} \mid \boldsymbol{s}, \boldsymbol{p})}{\pi^\mathsf{old}(o_{n} \mid \boldsymbol{s}, \boldsymbol{p})}, 1-\epsilon, 1+\epsilon\bigr) \cdot A_{n}\Bigr) \ ,
\end{align}
where $\pi^\mathsf{old}(\cdot \mid \cdot, \cdot)$ denotes the old policy, $v_n$ is the advantage associated with the prediction $o_n$, $\text{clip}(\cdot,\cdot,\cdot)$ is a clip function, and $\epsilon$ is a hyperparameter.
The advantage serves as a weight, encouraging high reward predictions while suppressing low reward ones.

\begin{table}[!t]
    \centering
    \caption{
    Comparison with state-of-the-art methods on  the ScreenSpot-Pro dataset \cite{li2025screenspotpro}. We report the grounding accuracy (\%) across various task domains, categorizing results by grounding target type: Text, Icon, and the overall average (Avg). We use `-' to denote unavailability, and `$^*$' to denote the results evaluated by us (which will be updated if improved evaluation scripts become available). The final average scores are highlighted in \textcolor{salesforcedarkblue!30!white}{dark blue}, and the best scores are in bold.
    } 
    \setlength{\tabcolsep}{1.2pt}
    \small
    \vspace{-1em}
    \begin{tabular}{@{}l ccc ccc ccc ccc ccc ccc cc>{\columncolor{salesforcedarkblue!30!white}}c @{}}
        \toprule
        \multirow{2}{*}{\textbf{Agent Model}} & \multicolumn{3}{c}{\textbf{Development}} & \multicolumn{3}{c}{\textbf{Creative}} & \multicolumn{3}{c}{\textbf{CAD}} & \multicolumn{3}{c}{\textbf{Scientific}} & \multicolumn{3}{c}{\textbf{Office}} & \multicolumn{3}{c}{\textbf{OS}} & \multicolumn{3}{c}{\textbf{Avg}} \\
        \cmidrule(lr){2-4} \cmidrule(lr){5-7} \cmidrule(lr){8-10} \cmidrule(lr){11-13} \cmidrule(lr){14-16} \cmidrule(lr){17-19} \cmidrule(lr){20-22}
        & Text & Icon & Avg & Text & Icon & Avg & Text & Icon & Avg & Text & Icon & Avg & Text & Icon & Avg & Text & Icon & Avg & Text & Icon & \cellcolor{white}{Avg} \\
        \midrule
        \rowcolor{gray!15}
        \multicolumn{22}{l}{\textit{Proprietary Models}} \\
        GPT-4o \cite{hurst2024gpt} & 1.3 & 0.0 & 0.7 & 1.0 & 0.0 & 0.6 & 2.0 & 0.0 & 1.5 & 2.1 & 0.0 & 1.2 & 1.1 & 0.0 & 0.9 & 0.0 & 0.0 & 0.0 & 1.3 & 0.0 & 0.8 \\
        Claude 3.7 Sonnet \cite{claude37} & - & - &- &- &- &- &- &- &- &- &- &- &- &- &- &- &- &- & - & - & 27.7 \\
        Operator~\cite{cua2025} & 50.0 & 19.3 & 35.1 & 51.5 & 23.1 & 39.6 & 16.8 & 14.1 & 16.1 & 58.3 & 24.5 & 43.7 & 60.5 & 28.3 & 53.0 & 34.6 & 30.3 & 32.7 & 45.0 & 23.0 & 36.6 \\
        Seed-1.5-VL \cite{guo2025seed1}  & -   & -   & 53.8	 & -   & -   & 59.2	  & -   & -   & 59.0 & -   & -   &	61.4  & -   & -   &	74.8  & -   & -   &	60.2 & -   & -   &	60.9 \\
        UI-TARS-1.5~\cite{qin2025ui} & -   & -   & 63.9 & -   & -   & 50.4	& -   & -   & 58.2	& -   & -   & 69.3 & -   & -   &	79.6 & -   & -   &	51.0 & -   & -   &	61.6 \\
        \midrule
        \rowcolor{gray!15}
        \multicolumn{22}{l}{\textit{Open-Source Models}} \\
        SeeClick~\cite{cheng2024seeclick} & 0.6 & 0.0 & 0.3 & 1.0 & 0.0 & 0.6 & 2.5 & 0.0 & 1.9 & 3.5 & 0.0 & 2.0 & 1.1 & 0.0 & 0.9 & 2.8 & 0.0 & 1.5 & 1.8 & 0.0 & 1.1 \\
        Qwen2-VL-7B~\cite{bai2025qwen2} & 2.6 & 0.0 & 1.3 & 1.5 & 0.0 & 0.9 & 0.5 & 0.0 & 0.4 & 6.3 & 0.0 & 3.5 & 3.4 & 1.9 & 3.0 & 0.9 & 0.0 & 0.5 & 2.5 & 0.2 & 1.6 \\
        ShowUI-2B~\cite{lin2025showui} & 16.9 & 1.4 & 9.4 & 9.1 & 0.0 & 5.3 & 2.5 & 0.0 & 1.9 & 13.2 & 7.3 & 10.6 & 15.3 & 7.5 & 13.5 & 10.3 & 2.2 & 6.6 & 10.8 & 2.6 & 7.7 \\
        CogAgent-18B~\cite{hong2024cogagent} & 14.9 & 0.7 & 8.0 & 9.6 & 0.0 & 5.6 & 7.1 & 3.1 & 6.1 & 22.2 & 1.8 & 13.4 & 13.0 & 0.0 & 10.0 & 5.6 & 0.0 & 3.1 & 12.0 & 0.8 & 7.7 \\
        Aria-UI~\cite{yang2024aria} & 16.2 & 0.0 & 8.4 & 23.7 & 2.1 & 14.7 & 7.6 & 1.6 & 6.1 & 27.1 & 6.4 & 18.1 & 20.3 & 1.9 & 16.1 & 4.7 & 0.0 & 2.6 & 17.1 & 2.0 & 11.3 \\
        UI-R1-3B \cite{lu2025ui} & 22.7 & 4.1 & - & 27.3 & 3.5 & - & 11.2 & 6.3 & - & 42.4 & 11.8 & - & 32.2 & 11.3 & - & 13.1 & 4.5 & - & - & - & 17.8 \\
        OS-Atlas-7B~\cite{wu2024atlas}  & 33.1 & 1.4 & 17.7 & 28.8 & 2.8 & 17.9 & 12.2 & 4.7 & 10.3 & 37.5 & 7.3 & 24.4 & 33.9 & 5.7 & 27.4 & 27.1 & 4.5 & 16.8 & 28.1 & 4.0 & 18.9 \\
        UI-TARS-2B~\cite{qin2023chatgpt} & 47.4 & 4.1 & 26.4 & 42.9 & 6.3 & 27.6 & 17.8 & 4.7 & 14.6 & 56.9 & 17.3 & 39.8 & 50.3 & 17.0 & 42.6 & 21.5 & 5.6 & 14.3 & 39.6 & 8.4 & 27.7 \\
        Qwen2.5-VL-3B \cite{bai2025qwen2} & 38.3 & 3.4 & 21.4 & 40.9 & 4.9 & 25.8 & 22.3 & 6.3 & 18.4 & 44.4 & 10.0 & 29.5 & 48.0 & 17.0 & 40.9 & 33.6 & 4.5 & 20.4 & 37.8 & 6.6 & 25.9 \\
        Qwen2.5-VL-7B  \cite{bai2025qwen2} & 51.9 & 4.8 & 29.1 & 36.9 & 8.4 & 24.9 & 17.8 & 1.6 & 13.8 & 48.6 & 8.2 & 31.1 & 53.7 & 18.9 & 45.7 & 34.6 & 7.9 & 22.4 & 39.9 & 7.6 & 27.6 \\
        UGround-7B~\cite{gou2024navigating} & -   & -   & 35.5 & -   & -   & 27.8 & -   & -   & 13.5 & -   & -   & 38.8 & -   & -   & 48.8 & -   & -   & 26.1 & -   & -   & 31.1 \\
        UGround-72B ~\cite{gou2024navigating}  & -   & -   & 31.1	 & -   & -   & 35.8  & -   & -   &	13.8  & -   & -   &	50.0	 & -   & -   & 51.3	  & -   & -   & 25.5	 & -   & -   & 34.5 \\
        UI-TARS-7B~\cite{qin2023chatgpt} & 58.4 & 12.4& 36.1 & 50.0 & 9.1 & 32.8 & 20.8 & 9.4 & 18.0 & 63.9 & 31.8 & 50.0 & 63.3 & 20.8 & 53.5 & 30.8 & 16.9 & 24.5 & 47.8 & 16.2 & 35.7 \\
        InfiGUI-R1-3B \cite{liu2025infigui} & 51.3  & 12.4 & 32.4 & 44.9 & 7.0 & 29.0 & 33.0 & 14.1 & 28.4 & 58.3 & 20.0 & 41.7 & 65.5 & 28.3 & 57.0 & 43.9 & 12.4 & 29.6 & 49.1 & 14.1 & 35.7 \\
        SE-GUI-3B~\cite{yuan2025enhancing} & 55.8 & 7.6 & 35.1 & 47.0 & 4.9 & 29.0 & 38.1 & 12.5 & 31.8 & 61.8 & 16.4 & 43.3 & 59.9 & 24.5 & 50.9 & 40.2 &  12.4 & 25.5 & 50.4 & 11.8 & 35.9 \\
        Jedi-3B \cite{xie2025scaling} & 61.0 & 13.8 & 38.1 & 53.5 & 8.4 & 34.6 & 27.4 & 9.4 & 23.0 & 54.2 & 18.2 & 38.6 & 64.4 & 32.1 & 57.0 & 38.3 & 9.0 & 25.0 & 49.8 & 13.7 & 36.1 \\
        GUI-G1-3B \cite{zhou2025gui}  & 50.7 & 10.3 & 31.1 & 36.6 & 11.9 & 26.6 & 39.6 & 9.4  & 32.2 & 61.8 & 30.0 & 48.0 & 67.2 & 32.1 & 59.1 & 23.5 & 10.6 & 16.1 & 49.5 & 16.8 & 37.1 \\
        UI-TARS-72B~\cite{qin2023chatgpt} & 63.0 & 17.3 & 40.8 & 57.1 & 15.4 & 39.6 & 18.8 & 12.5 & 17.2 & 64.6 & 20.9 & 45.7 & 63.3 & 26.4 & 54.8 & 42.1 & 15.7 & 30.1 & 50.9 & 17.5 & 38.1 \\
        Jedi-7B \cite{xie2025scaling} & 42.9 & 11.0 & 27.4 & 50.0 & 11.9 & 34.0 & 38.0 & 14.1 & 32.2 & 72.9 & 25.5 & 52.4 & 75.1 & 47.2 & 68.7 & 33.6 & 16.9 & 26.0 & 52.6 & 18.2 & 39.5 \\
        UI-TARS-1.5-7B$^*$~\cite{qin2023chatgpt} &  58.4 &  12.4 & 31.8 & 58.1 & 15.4 & 40.2 & 38.6 & 11.0 & 31.8 & 66.7 & 21.9 & 47.2 &74.6 & 35.9 & 65.6 & 49.5 & 13.5 & 33.2 & 57.5 & 16.9 & 42.0 \\
        Qwen2.5-VL-32B  \cite{bai2025qwen2} & 74.0 & 21.4 & 48.5 & 61.1 & 13.3 & 41.1 & 38.1 & 15.6 & 32.6 & 78.5 & 29.1 & 57.1 & 76.3 & 37.7 & 67.4 & 55.1 & 27.0 & 42.3 & 63.2 & 22.5 & 47.6 \\
        SE-GUI-7B~\cite{yuan2025enhancing} &  68.2 & 19.3 & 44.5 & 57.6 & 9.1 & 37.2 & 51.3 & \textbf{42.2} & 42.1 & 75.0 & 28.2 & 54.7 & 78.5 & 43.4 & 70.4 & 49.5 & 25.8 & 38.8 & 63.5 & 21.0 & 47.3 \\
        Qwen2.5-VL-72B \cite{bai2025qwen2} & -   & -   & 53.5	& -   & -   & 44.9	& -   & -   & 44.4	& -   & -   & 59.1 & -   & -   &	72.6& -   & -   &	49.5	& -   & -   & 53.3\\
        OpenCUA-32B \cite{wang2025opencuaopenfoundationscomputeruse} & - & - & - & - & - & - & - & - & - & - & - & - & - & -& - & - & -& - & - & - & 55.3 \\
        \midrule
        \name-7B & 66.9   & 20.7   & 44.5	& 62.6   & 18.2   & 44.0	& 53.3   & 17.2   & 44.4   & 76.4 & 31.8   &	57.1 & 82.5   & 50.9   &	75.2 & 48.6  & 25.9   &	38.3 & 65.5   & 25.2   &	50.1 \\
        \name-32B & \textbf{83.1} & \textbf{37.9} & \textbf{61.2} & \textbf{72.2} & \textbf{25.9} & \textbf{52.8} & \textbf{70.1} & 31.3 & \textbf{60.5} & \textbf{84.7} & \textbf{39.1} & \textbf{65.0} & \textbf{89.3} & \textbf{64.2} & \textbf{83.5} & \textbf{76.6} & \textbf{51.7} & \textbf{65.3} & \textbf{78.9} & \textbf{38.9} & \textbf{63.6} \\
        \bottomrule
    \end{tabular}
    \label{tab:screenspot_pro_comparison}
\end{table}

\section{Experiment}

\paragraph{Implementation Detail.} We train our model using a mixture of dataset \cite{yang2024aria, wu2024atlas, nayak2025uivisiondesktopcentricguibenchmark,kapoor2024omniactdatasetbenchmarkenabling, li2020widgetcaptioninggeneratingnatural}, applying a data cleaning threshold of $\tau = 0.3$. Our model is initialized from \cite{qin2025ui, wang2025opencuaopenfoundationscomputeruse}. When testing on the real-world dynamic environment, we use the action space from \cite{agashe2025agent} and o3 as the default planner \cite{openAI_o3_o4_mini}. Refer to \cref{app:training} for more details.

\paragraph{Dataset.} We evaluate our method on two sets of benchmarks: i) GUI Grounding, where we use ScreenSpot-Pro \cite{li2025screenspotpro}, ScreenSpot-V2 \cite{cheng2024seeclick,wu2024atlas}, and OSWorld-G \cite{xie2024osworld} datasets, evaluating by the metric of accuracy; ii) Agent Task Execution, where we use OSWorld \cite{xie2024osworld} and WindowsAgentArena \cite{bonatti2024windowsagentarenaevaluating} benchmarks, measuring performance by task success rate. For the OSWorld benchmark, we explore both its original release and OSWorld-Verified (\ie, an updated variant).

\paragraph{Baseline.} We compare with various state-of-the-art methods: CogAgent \cite{hong2024cogagent}, OminiParser \cite{lu2024omniparser}, Qwen2.5-VL \cite{bai2025qwen2}, Aria-UI \cite{yang2024aria}, OS-Atlas \cite{wu2024atlas}, UGround \cite{gou2024navigating}, ShowUI \cite{lin2025showui},  Aguvis \cite{xu2024aguvis}, Jedi \cite{xie2025scaling}, GUI-G1 \cite{zhou2025gui}, SE-GUI \cite{yuan2025enhancing}, GUI-R1 \cite{luo2025gui}, UI-R1 \cite{lu2025ui}, InfiGUI-R1 \cite{liu2025infigui}, UI-TARS \cite{qin2025ui}, Seed-1.5-VL \cite{guo2025seed1}, UI-TARS-1.5 \cite{qin2025ui}, GPT-4o \cite{hurst2024gpt}, 
o3 \cite{openAI_o3_o4_mini}, Claude 3.7 Sonnet \cite{claude37}, and Gemini-2.5 \cite{gemini25}.

\begin{table}[!t]
    \caption{Comparison with state-of-the-art methods on the ScreenSpot-V2  dataset \cite{cheng2024seeclick,wu2024atlas} across mobile, desktop, and web domains. We report grounding accuracy (\%) categorized by grounding target type: Text, Icon/Widget, and the overall Average (Avg). We use `-' to denote unavailability, and `$^*$' to denote the results evaluated by us (which will be updated if improved evaluation scripts become available). The final average scores are highlighted in \textcolor{salesforcedarkblue!30!white}{dark blue}. The best scores are in bold.}
    \vspace{-1em}
    \centering
    \setlength{\tabcolsep}{11pt}
    \small
    \begin{tabular}{lcccccc >{\columncolor{salesforcedarkblue!30!white}}c}
        \toprule
        \multirow{2}{*}{\textbf{Agent Model}} & \multicolumn{2}{c}{\textbf{Mobile}} & \multicolumn{2}{c}{\textbf{Desktop}} & \multicolumn{2}{c}{\textbf{Web}} & \multirow{2}{*}{\cellcolor{white}\textbf{Avg}} \\
        \cmidrule(lr){2-3} \cmidrule(lr){4-6} \cmidrule(lr){6-7} 
         & \textbf{Text} & \textbf{Icon/Widget} & \textbf{Text} & \textbf{Icon/Widget} & \textbf{Text} & \textbf{Icon/Widget} & \cellcolor{white}{} \\
        \midrule
        \rowcolor{gray!15}
        \multicolumn{8}{l}{\textit{Proprietary Models}} \\
        Operator \cite{cua2025} & 47.3 & 41.5  & 90.2 & 80.3 & 92.8 & 84.3 & 70.5 \\
        Claude 3.7 Sonnet \cite{claude37} & - & - & - & - & - & - &  87.6 \\
        UI-TARS-1.5 \cite{qin2025ui} & - & - & - & - & - & - &  94.2  \\
        Seed-1.5-VL \cite{guo2025seed1}  & - & - & - & - & - & - &  \textbf{95.2} \\
        \midrule
        \rowcolor{gray!15}
        \multicolumn{8}{l}{\textit{Open-Source Models}} \\
        SeeClick \cite{guo2025seed1} & 78.4 & 50.7 & 70.1 & 29.3 & 55.2 & 32.5 & 55.1 \\
        OmniParser-v2 \cite{lu2024omniparser} & 95.5 & 74.6 & 92.3 & 60.9 & 88.0 & 59.6 & 80.7 \\
        Qwen2.5-VL-3B \cite{bai2025qwen2} & 93.4 & 73.5 & 88.1 & 58.6 & 88.0 & 71.4 & 80.9 \\
        UI-TARS-2B \cite{qin2025ui} & 95.2 & 79.1 & 90.7 &  68.6 &  87.2 & 78.3 & 84.7 \\
        OS-Atlas-Base-7B \cite{wu2024atlas} & 95.2 & 75.8 & 90.7 & 63.6 & 90.6 & 77.3 & 85.1 \\
        OS-Atlas-Base-7B \cite{wu2024atlas} & 96.2 & 83.4 & 89.7 & 69.3 & 94.0 & 79.8 & 87.1 \\
        Jedi-3B \cite{xie2025scaling} & 96.6 & 81.5 & 96.9 & 78.6 & 88.5 & 83.7 & 88.6 \\
        Qwen2.5-VL-7B \cite{bai2025qwen2} & 97.6 & 87.2 & 90.2 & 74.2 & 93.2 & 81.3 & 88.8 \\
        UI-TARS-1.5-7B$^{*}$ \cite{qin2025ui} & 95.9 & 84.8 & 94.9 & 80.7 & 90.6 & 86.2 & 89.7 \\
        UI-TARS-72B \cite{qin2025ui} & 94.8 & 86.3 & 91.2 & 87.9 & 91.5 & 87.7 & 90.3 \\
        UI-TARS-7B \cite{qin2025ui} & 96.9 & 89.1 & 95.4 & 85.0 & 93.6 & 85.2 & 91.6 \\
        Jedi-7B \cite{xie2025scaling} & 96.9 & 87.2 & 95.9 & 87.9 & 94.4 & 84.2 & 91.7 \\
        Qwen2.5-VL-32B$^*$ \cite{bai2025qwen2} & 98.3 & 86.7 & 94.3 &  83.6 & 93.6 & 89.7 & 91.9 \\
        Qwen2.5-VL-72B$^*$ \cite{bai2025qwen2} & 99.0 & 90.1 & 96.4 & 87.1 & \textbf{96.6} & \textbf{90.6} & 94.0 \\
        OpenCUA-32B \cite{wang2025opencuaopenfoundationscomputeruse} & - & - & - & - & - & - & 93.4 \\
        \midrule
        \name-7B & 99.0 & 88.6 & 94.9 & 89.3 & 92.3 & 86.7 & 92.4  \\
        \name-32B & \textbf{99.7} & \textbf{90.5} & \textbf{99.0} & \textbf{94.3} & 95.7 & 90.1 & \textbf{95.2} \\
        \bottomrule
    \end{tabular}
    \label{tab:screenspot_v2_comparison}
\end{table}

\begin{table}[!t]
    \caption{Performance comparison of state-of-the-art models on the OSWorld-G \cite{xie2025scaling} dataset. We report grounding accuracy (\%) categorized by different capabilities, along with the overall average (Avg). We use `$^*$' to denote the results evaluated by us (which will be updated if improved evaluation scripts become available). For methods not tagged with $^\dagger$, the results are based on refined grounding instructions. The final average scores on the benchmark are highlighted in \textcolor{salesforcedarkblue!30!white}{dark blue}. The best scores are in bold. }
    \vspace{-1em}
    \centering
    \setlength{\tabcolsep}{10.5pt}
    \small
    \begin{tabular}{lccccc >{\columncolor{salesforcedarkblue!30!white}}c}
        \toprule
        \textbf{Agent Model} & \makecell[c]{\textbf{Text}\\\textbf{Matching}} & \makecell[c]{\textbf{Element}\\\textbf{Recognition}} & \makecell[c]{\textbf{Layout}\\\textbf{Understanding}} & \makecell[c]{\textbf{Fine-grained}\\\textbf{Manipulation}}& \textbf{Refusal} & \cellcolor{white}\textbf{Avg} \\
        \midrule
        \rowcolor{gray!15}
        \multicolumn{7}{l}{\textit{Proprietary Models}} \\
        Operator$^\dagger$  \cite{cua2025} & 51.3 & 42.4 & 46.6 & 31.5 & 0.0 & 40.6 \\
        Operator \cite{cua2025} & - & - & - & - & - & 57.8 \\
        Gemini-2.5-Pro \cite{gemini2} & 59.8 & 45.5 & 49.0 & 33.6 & \textbf{38.9} & 45.2 \\
        Gemini-2.5-Pro \cite{gemini2} & - & - & - & - & - & 57.5 \\
        Seed1.5-VL$^\dagger$ \cite{guo2025seed1} & 73.9 & 66.7 & 69.6 & 47.0 & 18.5 & 62.9 \\
        \midrule
        \rowcolor{gray!15}
        \multicolumn{7}{l}{\textit{Open-Source Models}} \\
        Qwen2.5-VL-3B$^\dagger$ \cite{bai2025qwen2}  & 41.4 & 28.8 & 34.8 & 13.4 & 0.0 & 27.3 \\
        OS-Atlas-7B$^\dagger$ \cite{wu2024atlas}& 44.1 & 29.4 & 35.2 & 16.8 & 7.4 & 27.7 \\
        Qwen2.5-VL-7B$^\dagger$  \cite{bai2025qwen2} & 45.6 & 32.7 & 41.9 & 18.1 & 0.0 & 31.4 \\
        UGround-7B$^\dagger$ \cite{gou2024navigating} & 51.3 & 40.3 & 43.5 & 24.8 & 0.0 & 36.4 \\
        Aguvis-7B$^\dagger$ \cite{xu2024aguvis} & 55.9 & 41.2 & 43.9 & 28.2 & 0.0 & 38.7 \\
        UI-TARS-7B$^\dagger$ \cite{qin2025ui} & 60.2 & 51.8 & 54.9 & 35.6 & 0.0 & 47.5 \\
        Qwen2.5-VL-32B  \cite{bai2025qwen2}& 57.9 &  70.2 & 73.8 & 49.2 & 0.0 & 59.6\\
        Jedi-3B$^\dagger$ \cite{xie2025scaling} & 67.4 & 53.0 & 53.8 & 44.3 & 7.4 & 50.9 \\
        Jedi-3B \cite{xie2025scaling} & - & - & - & - & - & 61.0 \\
        Jedi-7B$^\dagger$ \cite{xie2025scaling} & 65.9 & 55.5 & 57.7 & 46.9 & 7.4 & 54.1 \\
        Jedi-7B \cite{xie2025scaling} & - & - & - & - & - & 63.8 \\
        UI-TARS-72B$^\dagger$ \cite{qin2025ui}& \textbf{69.4} & 60.6 & 62.9 & 45.6 & 0.0 & 57.1 \\
        Qwen2.5-VL-72B  \cite{bai2025qwen2} & 52.6 & 74.6 & 74.7 & 55.3 & 0.0 & 62.2 \\
        UI-TARS-1.5-7B$^{*\dagger}$ \cite{qin2025ui} & 36.8 & 62.7 & 62.2 & 50.8 & 0.0 & 52.8\\
        UI-TARS-1.5-7B$^{*}$ \cite{qin2025ui} & 52.6 & 75.4 & 72.4 & 66.7 & 0.0 & 64.2\\
        OpenCUA-32B$^\dagger$ \cite{wang2025opencuaopenfoundationscomputeruse} & - & - & - & - & - & 59.6 \\
        OpenCUA-32B$^{*}$ \cite{wang2025opencuaopenfoundationscomputeruse} & 63.2 & 79.9 & \textbf{84.9} & 62.1 & 7.4 & 70.2 \\
        \midrule
        \name-7B$^\dagger$ & 42.1 & 65.7 & 62.7 & 56.1 & 0.0 & 55.1 \\
        \name-7B & 63.2 & 82.1 & 74.2  & \textbf{70.5} & 0.0 & 67.7 \\
        \name-32B$^\dagger$ & 63.2 & 78.4 & 73.3 & 65.2 &0.0& 65.2  \\
        \name-32B & 63.2  &  \textbf{83.6} & 84.4 & \textbf{70.5} & 0.0 & \textbf{72.2} \\
        \bottomrule
    \end{tabular}
    \label{tab:osworld_g_comparison}
\end{table}

\begin{table}[!t]
\caption{Comparison with state-of-the-art methods on the OSWorld and OSWorld-Verified \cite{xie2024osworld} benchmarks. A dash (`-') indicates unavailable results, the second column shows the number of steps, and success rate success rate (\%) is reported. The best scores are in bold.} 
\setlength{\tabcolsep}{25pt}
\vspace{-1em}
\small
    \centering
    \begin{tabular}{lccc}
        \toprule
        \textbf{Agent Model} & \textbf{Step} & \textbf{OSWorld} &  \textbf{OSWorld-Verified}\\
        \midrule
        \rowcolor{gray!15}
        \multicolumn{4}{l}{\textit{Proprietary Models}} \\
        Claude 3.7 Sonnet \cite{claude37} & 100 & 28.0 & - \\
        OpenAI CUA 4o \cite{openAI_o3_o4_mini} & 200 & 38.1 & - \\
        UI-TARS-1.5  \cite{qin2025ui} & 100 & 42.5 & 41.8  \\
        OpenAI CUA o3 \cite{openAI_o3_o4_mini} & 200 & 42.9  & -  \\
        \midrule
        \rowcolor{gray!15}
        \multicolumn{4}{l}{\textit{Open-Source Models}} \\
        Aria-UI w/ GPT-4o~\cite{yang2024aria} & 15 & 15.2 & -\\
        Aguvis-72B w/ GPT-4o~\cite{xu2024aguvis} & 15 & 17.0  & - \\
        UI-TARS-72B-SFT \cite{qin2025ui} & 50 & 18.8 & - \\
        Agent S w/ Claude-3.5-Sonnet ~\cite{agashe2024agent}  & 15 & 20.5 \\
        Agent S w/ GPT-4o~\cite{agashe2024agent}  & 15 & 20.6 & -   \\
        UI-TARS-72B-DPO \cite{qin2025ui} & 15 & 22.7 & -  \\
        UI-TARS-72B-DPO \cite{qin2025ui} & 50 & 24.6 & - \\
        UI-TARS-1.5-7B  \cite{qin2025ui} & 100 & 26.9 & 27.4 \\
        Jedi-7B w/ o3 \cite{xie2025scaling} & 100 & - & 51.0   \\
        Jedi-7B w/ GPT-4o \cite{xie2025scaling} & 100 & 27.0 & - \\
        Agent S2 w/ Claude-3.7-Sonnet \cite{agashe2025agent} & 50 & 34.5 & - \\
        Agent S2 w/ Gemini-2.5-Pro \cite{agashe2025agent} & 50 & 41.4  & 45.8\\
        Agent S2.5 w/ o3\cite{agashe2025agent} & 100 & - & 56.0 \\
        Agent S2.5 w/ GPT-5\cite{agashe2025agent} & 100 & - & 58.4 \\
        CoAct-1 w/o3 \& o4mini \& OpenAI CUA 4o \cite{song2025coact1computerusingagentscoding} & 150 & - & 60.8 \\
        \midrule
        \name-7B w/ o3  & 100 & \textbf{45.2} & 53.1 \\
        \name-7B w/ GPT-5 & 100 & - &  61.0 \\
        \name-32B w/ o3  & 100 & - & 55.4 \\
        \name-32B w/ GPT-5 & 100 & - & \textbf{63.4} \\
        \bottomrule
    \end{tabular}
    \label{tab:osworld_comparison}
\end{table}

\begin{table}[!t]
    \centering
    \caption{Comparisons on WindowsAgentArena \cite{bonatti2024windowsagentarenaevaluating} benchmarks. We report success rate (\%) for evaluations. The best scores are in bold.}
    \vspace{-1em}
     \small
    \label{tab:winarena_comparision}
    \setlength{\tabcolsep}{55pt}
    \begin{tabular}{lcc}
        \toprule
       \textbf{Agent Model}   &  \textbf{Step}  &  \textbf{Success Rate} \\
        \midrule
         Kimi-VL  \cite{kimiteam2025kimivltechnicalreport} & 15 & 10.4 \\
         WAA \cite{bonatti2024windowsagentarenaevaluating}     & - & 19.5 \\
         Jedi-7B w/ GPT-4o \cite{xie2025scaling} & 100 & 33.7 \\
         GTA1-7B w/ o3 & 100 & 47.9 \\
         GTA1-7B w/ GPT-5 & 100 & 49.2 \\
         GTA1-32B w/ o3 & 100 & \textbf{51.2} \\
         GTA1-32B w/ GPT-5 & 100 & 50.6 \\
         \bottomrule
    \end{tabular}
\end{table}

\subsection{Grounding Performance}
We compare our method with state-of-the-art approaches on the ScreenSpot-Pro \cite{li2025screenspotpro}, ScreenSpot-V2 \cite{guo2025deepseek,wu2024atlas}, and OSWorld-G \cite{xie2024osworld} datasets, as shown in \cref{tab:screenspot_pro_comparison}, \cref{tab:screenspot_v2_comparison}, and \cref{tab:osworld_g_comparison}, respectively. Among the three benchmarks, the ScreenSpot-Pro benchmark is the most challenging, designed for high-resolution, complex, and professional GUI grounding scenarios. The ScreenSpot-V2 benchmark evaluates grounding capability across mobile, desktop, and web domains, while the OSWorld-G benchmark focuses on the Linux environment, providing a comprehensive benchmark for measuring diverse capabilities such as text matching, element recognition, layout understanding, and precise manipulation.

Our method consistently demonstrates the best performance. On the ScreenSpot-Pro \cite{li2025screenspotpro} benchmark, our 7B model outperforms much larger alternatives, for example, achieving 50.1\% scores compared to 34.5\% scores from UGround-72B. On the ScreenSpot-V2 benchmark, our best-performing model, \name-32B, achieves the same performance as the proprietary Seed-1.5-VL \cite{guo2025seed1}. Similarly, on the OSWorld-G \cite{xie2025scaling} benchmark, our method surpasses all state-of-the-art approaches, setting a new benchmark with a grounding accuracy of 72.2\%.

Overall, our method achieves state-of-the-art performance using a simple click-based training strategy, demonstrating both robustness and effectiveness. This highlights its potential as a strong foundation for grounding models in complex GUI environments.

\subsection{Agent Performance}
We compare our method with state-of-the-art approaches on the OSWorld \cite{xie2024osworld} benchmark in \cref{tab:osworld_comparison}.  It consists of 369 tasks distributed across real-world web and desktop applications, providing a diverse and challenging testbed for assessing agent capabilities to complete user tasks in Linux environments. We assess various scales of our grounding model using o3 as the planner and the judge for our test-time scaling strategy, forming the GTA1 agent series. Among the three model scales, GTA1-7B achieves the highest task success rate of 45.2\% on the OSWorld benchmark, outperforming all state-of-the-art methods. It is worth highlighting that, even when using the o3 planner, GTA1-7B significantly outperforms its native agent variant, CUA o3, while operating with a shorter execution horizon (\ie, 45.2\% from our method with a 100-step horizon \textit{vs.} 42.9\% from CUA o3 with a 200-step horizon \cite{cua2025}).

The strong performance of the GTA1 agent demonstrates its effectiveness in handling complex and real-world user tasks across diverse scenarios. This highlights the robustness and generalization capability of our approach. In the following subsections, we present a more detailed analysis of the GTA1 agent to examine it. 

\begin{table}[!t]
    \centering
    \caption{Ablation of optimization rewards. We study three types of rewards used to guide model learning: click reward (\ie, whether the prediction falls within the target element bounding box), IoU reward (\ie, enforcing predictions of the bounding box of the target element), and format reward (\ie, enforcing ``thinking'' before predictions).
    }
    \vspace{-1em}
    \label{tab:ablation}
    \centering
    \setlength{\tabcolsep}{15pt}
    \begin{tabular}{cccccc}
        \toprule
        \makecell[c]{\bf Click\\\bf Reward} & \makecell[c]{\bf IoU\\ \bf Reward} & \makecell[c]{\bf Format\\\bf Reward} & \textbf{ScreenSpot-Pro} & \textbf{ScreenSpot-V2}  & \textbf{OSWorld-G} \\
        \midrule
        \checkmark & \checkmark & \checkmark & 44.5 & 89.3 & 59.9 \\
        \checkmark & \checkmark & &  42.2 & 89.2 & 59.2 \\
        \checkmark & & \checkmark & 46.9 &  93.2 & 67.0 \\
        \checkmark & & & 50.1 & 92.4 & 67.7\\
        \bottomrule
    \end{tabular}
\end{table}

\begin{figure}[!t]
\centering
\includegraphics{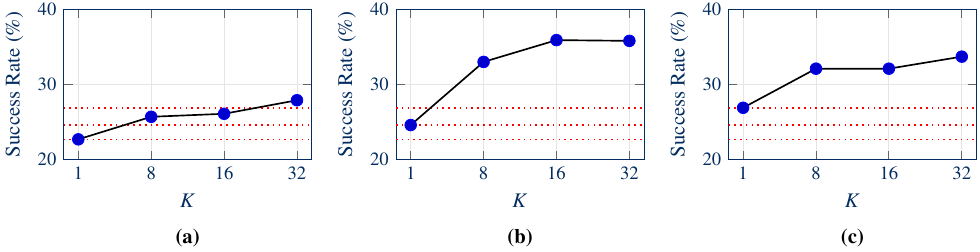}
\vspace{-2em}
\caption{Scalability of the number $K$ of action proposals in test-time scaling. The task execution steps in \textbf{(a)}, \textbf{(b)}, and \textbf{(c)} are 15, 50, and 100, respectively. We vary $K$ over \{1, 8, 16, 32\}, and measure performance using task success rate (\%). When $K = 1$, it degrades to the setting where the test-time scaling strategy is not used. We color the performance of $K = 1$ over \{15, 50, 100\} steps on the OSWorld benchmark with red dotted lines.
}
\label{fig:K}
\end{figure}

\subsection{Discussion and Ablation}

\paragraph{Click Reward Works the Best.} We explore different optimization objectives for \name, focusing on widely studied rewards such as the format reward (\ie, enforcing ``thinking'') and IoU rewards (\ie, encouraging accurate bounding box predictions for the target element) in \cref{tab:ablation}. We evaluate three settings: optimizing the model using the format reward, IoU reward, and click reward (\cref{eq:reward}); 
using the IoU reward and click reward; 
applying the format reward and click reward. 
The three settings achieve performance of 44.5\%/42.2\%/46.9\%, 89.3\%/89.2\%/93.2\%, and 59.9\%/59.2\%/67.0\%  respectively on the ScreenSpot-Pro \cite{li2025screenspotpro}, ScreenSpot-V2 \cite{cheng2024seeclick,wu2024atlas}, and OSWorld-G \cite{xie2024osworld} benchmarks. However, all of the settings generally underperform compared to using the click reward alone (except on the ScreenSpot-V2 benchmark), which yields the accuracies of 50.1\%, 92.4\%, and 67.7\% on the three benchmarks.

\paragraph{``thinking'' Benefits Grounding in Dynamic Environment Only.} Across various benchmarks, we observe minimal performance differences between grounding models trained with and without ``thinking''. However, they often succeed on different samples, likely due to training instability rather than systematic reasoning gains. 
We find that ``thinking'' can be effective in dynamic environments such as the AndroidWorld benchmark \cite{rawles2024androidworld}, where the model is provided with the task object, past trajectories, and the user instruction. For example, by training an in-domain 7B model based on  \cite{bai2025qwen2}  using the AndroidControl dataset \cite{li2024effects}, using or not using ``thinking'' have similar grounding performance on the AndroidControl test fold. However, the task success rate on the AndroidWorld benchmark increased from 39\% to 44\% when using ``thinking''. This improvement is attributed to the increased complexity of the textual prompts (\ie, combination of task object, past trajectories, and the user instruction), which encourages the model to engage in ``thinking'' when operating under challenging and dynamic conditions.

\paragraph{Test-time Scaling Generalizes Well.} We demonstrate the generalization capability of our test-time scaling strategies through two sets of experiments. First, with $K=8$ action proposals, increasing the horizon from 50 to 100 steps on the OSWorld benchmark \cite{xie2024osworld} improves the success rate from 43.4\% to 45.2\%, indicating robustness across varying execution lengths. In comparison, the baseline performance with $K=1$ is 41.3\% and 43.4\% for the 50-step and 100-step horizons, respectively. Second, we show that our test-time scaling strategies generalize to other agents, as evidenced by the scalability of $K$ on UI-TARS-1.5-7B \cite{qin2025ui}, shown in \cref{fig:K}. We assess performance with 15-, 50-, and 100-step horizons in \cref{fig:K}~(a),  \cref{fig:K}~(b), and \cref{fig:K}~(c) respectively, by varying $K$ over \{1, 8, 16, 32\}. The red dotted lines mark the baseline performance of UI-TARS-1.5-7B without any test-time scaling (\ie, $K=1$) at each horizon on the OSWorld benchmark. Our test-time scaling strategy consistently boosts performance, yielding two main insights:
i) with our test-time scaling, UI-TARS-1.5-7B executed for only 15 steps and $K=32$ already outperforms the baseline that executes for 100 steps without scaling. Since the $K$ candidate action proposals are sampled concurrently, this also cuts wall-clock time substantially;
ii) the greatest overall gain occurs with a 50-step horizon. Using 15-step horizon is occasionally insufficient to complete certain tasks, whereas 100-step horizon provide excessive slack, diluting the benefit of additional steps. 
Moreover, we present the qualitative comparisons of UI-TARS-1.5-7B with and without test-time scaling strategies in \cref{app:example}, highlighting the effectiveness of the test-time strategy in making the agent highly susceptible to cascading failures.

\section{Conclusion and Limitation}
This paper investigates two key challenges toward building intelligent GUI agents:  selecting effective plans and precise grounding in complex interfaces. We address these challenges with two strategies. First, to improve task planning, we introduce a scalable test-time strategy that concurrent samples multiple action proposals at each step and use a multimodal large language model judge to select the most suitable one.  Second, we propose a grounding model, which employs a simple RL-based optimization approach that directly rewards successful clicks on target elements, bypassing the explicit ``thinking'' required by prior methods.  Overall, \name achieves state-of-the-art performance on standard  grounding benchmarks and demonstrates robust behavior when integrated with a planner and our test-time scaling strategy for user task execution in GUI environment. 
However, our approach has limitations. 
Although it achieves the highest accuracy on the challenging ScreenSpot-Pro benchmark, it still struggles in certain scenarios. For example, similar to prior work, our grounding model has difficulty in selecting custom foregrounds and backgrounds when applied to image editing with GMIP. We hope this work inspires further research on GUI agents.

\bibliography{references}

\begin{thebibliography}{10}

\bibitem{yang2024aria}
Y.~Yang, Y.~Wang, D.~Li, Z.~Luo, B.~Chen, C.~Huang, and J.~Li, ``Aria-ui: Visual grounding for gui instructions,'' {\em arXiv preprint arXiv:2412.16256}, 2024.

\bibitem{gou2024navigating}
B.~Gou, R.~Wang, B.~Zheng, Y.~Xie, C.~Chang, Y.~Shu, H.~Sun, and Y.~Su, ``Navigating the digital world as humans do: Universal visual grounding for gui agents,'' {\em arXiv preprint arXiv:2410.05243}, 2024.

\bibitem{li2025screenspotpro}
K.~Li, Z.~Meng, H.~Lin, Z.~Luo, Y.~Tian, J.~Ma, Z.~Huang, and T.-S. Chua, ``Screenspot-pro: Gui grounding for professional high-resolution computer use,'' {\em arXiv preprint arXiv:2504.07981}, 2025.

\bibitem{wu2024atlas}
Z.~Wu, Z.~Wu, F.~Xu, Y.~Wang, Q.~Sun, C.~Jia, K.~Cheng, Z.~Ding, L.~Chen, P.~P. Liang, {\em et~al.}, ``Os-atlas: A foundation action model for generalist gui agents,'' {\em arXiv preprint arXiv:2410.23218}, 2024.

\bibitem{cheng2024seeclick}
K.~Cheng, Q.~Sun, Y.~Chu, F.~Xu, Y.~Li, J.~Zhang, and Z.~Wu, ``Seeclick: Harnessing gui grounding for advanced visual gui agents,'' {\em arXiv preprint arXiv:2401.10935}, 2024.

\bibitem{xie2025scaling}
T.~Xie, J.~Deng, X.~Li, J.~Yang, H.~Wu, J.~Chen, W.~Hu, X.~Wang, Y.~Xu, Z.~Wang, {\em et~al.}, ``Scaling computer-use grounding via user interface decomposition and synthesis,'' {\em arXiv preprint arXiv:2505.13227}, 2025.

\bibitem{agashe2024agent}
S.~Agashe, J.~Han, S.~Gan, J.~Yang, A.~Li, and X.~E. Wang, ``Agent s: An open agentic framework that uses computers like a human,'' {\em arXiv preprint arXiv:2410.08164}, 2024.

\bibitem{agashe2025agent}
S.~Agashe, K.~Wong, V.~Tu, J.~Yang, A.~Li, and X.~E. Wang, ``Agent s2: A compositional generalist-specialist framework for computer use agents,'' {\em arXiv preprint arXiv:2504.00906}, 2025.

\bibitem{xu2024aguvis}
Y.~Xu, Z.~Wang, J.~Wang, D.~Lu, T.~Xie, A.~Saha, D.~Sahoo, T.~Yu, and C.~Xiong, ``Aguvis: Unified pure vision agents for autonomous gui interaction,'' {\em arXiv preprint arXiv:2412.04454}, 2024.

\bibitem{openAI_o3_o4_mini}
OpenAI, ``Openai o3 and o4-mini system card,'' technical report, OpenAI, 2025.
\newblock System Card.

\bibitem{xie2024osworld}
T.~Xie, D.~Zhang, J.~Chen, X.~Li, S.~Zhao, R.~Cao, T.~J. Hua, Z.~Cheng, D.~Shin, F.~Lei, {\em et~al.}, ``Osworld: Benchmarking multimodal agents for open-ended tasks in real computer environments,'' {\em Advances in Neural Information Processing Systems}, vol.~37, pp.~52040--52094, 2024.

\bibitem{guo2025deepseek}
D.~Guo, D.~Yang, H.~Zhang, J.~Song, R.~Zhang, R.~Xu, Q.~Zhu, S.~Ma, P.~Wang, X.~Bi, {\em et~al.}, ``Deepseek-r1: Incentivizing reasoning capability in llms via reinforcement learning,'' {\em arXiv preprint arXiv:2501.12948}, 2025.

\bibitem{shao2024deepseekmath}
Z.~Shao, P.~Wang, Q.~Zhu, R.~Xu, J.~Song, X.~Bi, H.~Zhang, M.~Zhang, Y.~Li, Y.~Wu, {\em et~al.}, ``Deepseekmath: Pushing the limits of mathematical reasoning in open language models,'' {\em arXiv preprint arXiv:2402.03300}, 2024.

\bibitem{luo2025gui}
R.~Luo, L.~Wang, W.~He, and X.~Xia, ``Gui-r1: A generalist r1-style vision-language action model for gui agents,'' {\em arXiv preprint arXiv:2504.10458}, 2025.

\bibitem{lu2025ui}
Z.~Lu, Y.~Chai, Y.~Guo, X.~Yin, L.~Liu, H.~Wang, H.~Xiao, S.~Ren, G.~Xiong, and H.~Li, ``Ui-r1: Enhancing action prediction of gui agents by reinforcement learning,'' {\em arXiv preprint arXiv:2503.21620}, 2025.

\bibitem{liu2025infigui}
Y.~Liu, P.~Li, C.~Xie, X.~Hu, X.~Han, S.~Zhang, H.~Yang, and F.~Wu, ``Infigui-r1: Advancing multimodal gui agents from reactive actors to deliberative reasoners,'' {\em arXiv preprint arXiv:2504.14239}, 2025.

\bibitem{zhou2025gui}
Y.~Zhou, S.~Dai, S.~Wang, K.~Zhou, Q.~Jia, {\em et~al.}, ``Gui-g1: Understanding r1-zero-like training for visual grounding in gui agents,'' {\em arXiv preprint arXiv:2505.15810}, 2025.

\bibitem{hurst2024gpt}
A.~Hurst, A.~Lerer, A.~P. Goucher, A.~Perelman, A.~Ramesh, A.~Clark, A.~Ostrow, A.~Welihinda, A.~Hayes, A.~Radford, {\em et~al.}, ``Gpt-4o system card,'' {\em arXiv preprint arXiv:2410.21276}, 2024.

\bibitem{claude37}
Anthropic, ``Claude 3.7 sonnet and claude code,'' technical report, Anthropic, 2025.
\newblock System Card.

\bibitem{qin2025ui}
Y.~Qin, Y.~Ye, J.~Fang, H.~Wang, S.~Liang, S.~Tian, J.~Zhang, J.~Li, Y.~Li, S.~Huang, {\em et~al.}, ``Ui-tars: Pioneering automated gui interaction with native agents,'' {\em arXiv preprint arXiv:2501.12326}, 2025.

\bibitem{lu2024omniparser}
Y.~Lu, J.~Yang, Y.~Shen, and A.~Awadallah, ``Omniparser for pure vision based gui agent,'' {\em arXiv preprint arXiv:2408.00203}, 2024.

\bibitem{cua2025}
OpenAI, ``Computer-using agent: Introducing a universal interface for ai to interact with the digital world,'' 2025.

\bibitem{guo2025seed1}
D.~Guo, F.~Wu, F.~Zhu, F.~Leng, G.~Shi, H.~Chen, H.~Fan, J.~Wang, J.~Jiang, J.~Wang, {\em et~al.}, ``Seed1. 5-vl technical report,'' {\em arXiv preprint arXiv:2505.07062}, 2025.

\bibitem{bai2025qwen2}
S.~Bai, K.~Chen, X.~Liu, J.~Wang, W.~Ge, S.~Song, K.~Dang, P.~Wang, S.~Wang, J.~Tang, {\em et~al.}, ``Qwen2. 5-vl technical report,'' {\em arXiv preprint arXiv:2502.13923}, 2025.

\bibitem{lin2025showui}
K.~Q. Lin, L.~Li, D.~Gao, Z.~Yang, S.~Wu, Z.~Bai, S.~W. Lei, L.~Wang, and M.~Z. Shou, ``Showui: One vision-language-action model for gui visual agent,'' in {\em Proceedings of the Computer Vision and Pattern Recognition Conference}, pp.~19498--19508, 2025.

\bibitem{hong2024cogagent}
W.~Hong, W.~Wang, Q.~Lv, J.~Xu, W.~Yu, J.~Ji, Y.~Wang, Z.~Wang, Y.~Dong, M.~Ding, {\em et~al.}, ``Cogagent: A visual language model for gui agents,'' in {\em Proceedings of the IEEE/CVF Conference on Computer Vision and Pattern Recognition}, pp.~14281--14290, 2024.

\bibitem{qin2023chatgpt}
C.~Qin, A.~Zhang, Z.~Zhang, J.~Chen, M.~Yasunaga, and D.~Yang, ``Is chatgpt a general-purpose natural language processing task solver?,'' {\em arXiv preprint arXiv:2302.06476}, 2023.

\bibitem{yuan2025enhancing}
X.~Yuan, J.~Zhang, K.~Li, Z.~Cai, L.~Yao, J.~Chen, E.~Wang, Q.~Hou, J.~Chen, P.-T. Jiang, {\em et~al.}, ``Enhancing visual grounding for gui agents via self-evolutionary reinforcement learning,'' {\em arXiv preprint arXiv:2505.12370}, 2025.

\bibitem{wang2025opencuaopenfoundationscomputeruse}
X.~Wang, B.~Wang, D.~Lu, J.~Yang, T.~Xie, J.~Wang, J.~Deng, X.~Guo, Y.~Xu, C.~H. Wu, Z.~Shen, Z.~Li, R.~Li, X.~Li, J.~Chen, B.~Zheng, P.~Li, F.~Lei, R.~Cao, Y.~Fu, D.~Shin, M.~Shin, J.~Hu, Y.~Wang, J.~Chen, Y.~Ye, D.~Zhang, D.~Du, H.~Hu, H.~Chen, Z.~Zhou, H.~Yao, Z.~Chen, Q.~Gu, Y.~Wang, H.~Wang, D.~Yang, V.~Zhong, F.~Sung, Y.~Charles, Z.~Yang, and T.~Yu, ``Opencua: Open foundations for computer-use agents,'' 2025.

\bibitem{nayak2025uivisiondesktopcentricguibenchmark}
S.~Nayak, X.~Jian, K.~Q. Lin, J.~A. Rodriguez, M.~Kalsi, R.~Awal, N.~Chapados, M.~T. Özsu, A.~Agrawal, D.~Vazquez, C.~Pal, P.~Taslakian, S.~Gella, and S.~Rajeswar, ``Ui-vision: A desktop-centric gui benchmark for visual perception and interaction,'' 2025.

\bibitem{kapoor2024omniactdatasetbenchmarkenabling}
R.~Kapoor, Y.~P. Butala, M.~Russak, J.~Y. Koh, K.~Kamble, W.~Alshikh, and R.~Salakhutdinov, ``Omniact: A dataset and benchmark for enabling multimodal generalist autonomous agents for desktop and web,'' 2024.

\bibitem{li2020widgetcaptioninggeneratingnatural}
Y.~Li, G.~Li, L.~He, J.~Zheng, H.~Li, and Z.~Guan, ``Widget captioning: Generating natural language description for mobile user interface elements,'' 2020.

\bibitem{bonatti2024windowsagentarenaevaluating}
R.~Bonatti, D.~Zhao, F.~Bonacci, D.~Dupont, S.~Abdali, Y.~Li, Y.~Lu, J.~Wagle, K.~Koishida, A.~Bucker, L.~Jang, and Z.~Hui, ``Windows agent arena: Evaluating multi-modal os agents at scale,'' 2024.

\bibitem{gemini25}
Deepmind, ``Gemini 2.5: Our most intelligent ai model,'' technical report, Deepmind, 2025.

\bibitem{gemini2}
Deepmind, ``Introducing gemini 2.0: our new ai model for the agentic era,'' technical report, Deepmind, 2025.

\bibitem{song2025coact1computerusingagentscoding}
L.~Song, Y.~Dai, V.~Prabhu, J.~Zhang, T.~Shi, L.~Li, J.~Li, S.~Savarese, Z.~Chen, J.~Zhao, R.~Xu, and C.~Xiong, ``Coact-1: Computer-using agents with coding as actions,'' 2025.

\bibitem{kimiteam2025kimivltechnicalreport}
K.~Team, A.~Du, B.~Yin, B.~Xing, B.~Qu, B.~Wang, C.~Chen, C.~Zhang, C.~Du, C.~Wei, C.~Wang, D.~Zhang, D.~Du, D.~Wang, E.~Yuan, E.~Lu, F.~Li, F.~Sung, G.~Wei, G.~Lai, H.~Zhu, H.~Ding, H.~Hu, H.~Yang, H.~Zhang, H.~Wu, H.~Yao, H.~Lu, H.~Wang, H.~Gao, H.~Zheng, J.~Li, J.~Su, J.~Wang, J.~Deng, J.~Qiu, J.~Xie, J.~Wang, J.~Liu, J.~Yan, K.~Ouyang, L.~Chen, L.~Sui, L.~Yu, M.~Dong, M.~Dong, N.~Xu, P.~Cheng, Q.~Gu, R.~Zhou, S.~Liu, S.~Cao, T.~Yu, T.~Song, T.~Bai, W.~Song, W.~He, W.~Huang, W.~Xu, X.~Yuan, X.~Yao, X.~Wu, X.~Li, X.~Zu, X.~Zhou, X.~Wang, Y.~Charles, Y.~Zhong, Y.~Li, Y.~Hu, Y.~Chen, Y.~Wang, Y.~Liu, Y.~Miao, Y.~Qin, Y.~Chen, Y.~Bao, Y.~Wang, Y.~Kang, Y.~Liu, Y.~Dong, Y.~Du, Y.~Wu, Y.~Wang, Y.~Yan, Z.~Zhou, Z.~Li, Z.~Jiang, Z.~Zhang, Z.~Yang, Z.~Huang, Z.~Huang, Z.~Zhao, Z.~Chen, and Z.~Lin, ``Kimi-vl technical report,'' 2025.

\bibitem{rawles2024androidworld}
C.~Rawles, S.~Clinckemaillie, Y.~Chang, J.~Waltz, G.~Lau, M.~Fair, A.~Li, W.~Bishop, W.~Li, F.~Campbell-Ajala, {\em et~al.}, ``Androidworld: A dynamic benchmarking environment for autonomous agents,'' {\em arXiv preprint arXiv:2405.14573}, 2024.

\bibitem{li2024effects}
W.~Li, W.~E. Bishop, A.~Li, C.~Rawles, F.~Campbell-Ajala, D.~Tyamagundlu, and O.~Riva, ``On the effects of data scale on ui control agents,'' {\em Advances in Neural Information Processing Systems}, vol.~37, pp.~92130--92154, 2024.

\bibitem{kwon2023efficientmemorymanagementlarge}
W.~Kwon, Z.~Li, S.~Zhuang, Y.~Sheng, L.~Zheng, C.~H. Yu, J.~E. Gonzalez, H.~Zhang, and I.~Stoica, ``Efficient memory management for large language model serving with pagedattention,'' 2023.

\end{thebibliography}
\bibliographystyle{ieeetr}

\newpage
\DoToC
\newpage
\appendix
\section{Example}
\label{app:example}
\begin{figure}[p]
    \centering
    \vspace{-1.5em}
    \includegraphics[width=\linewidth]{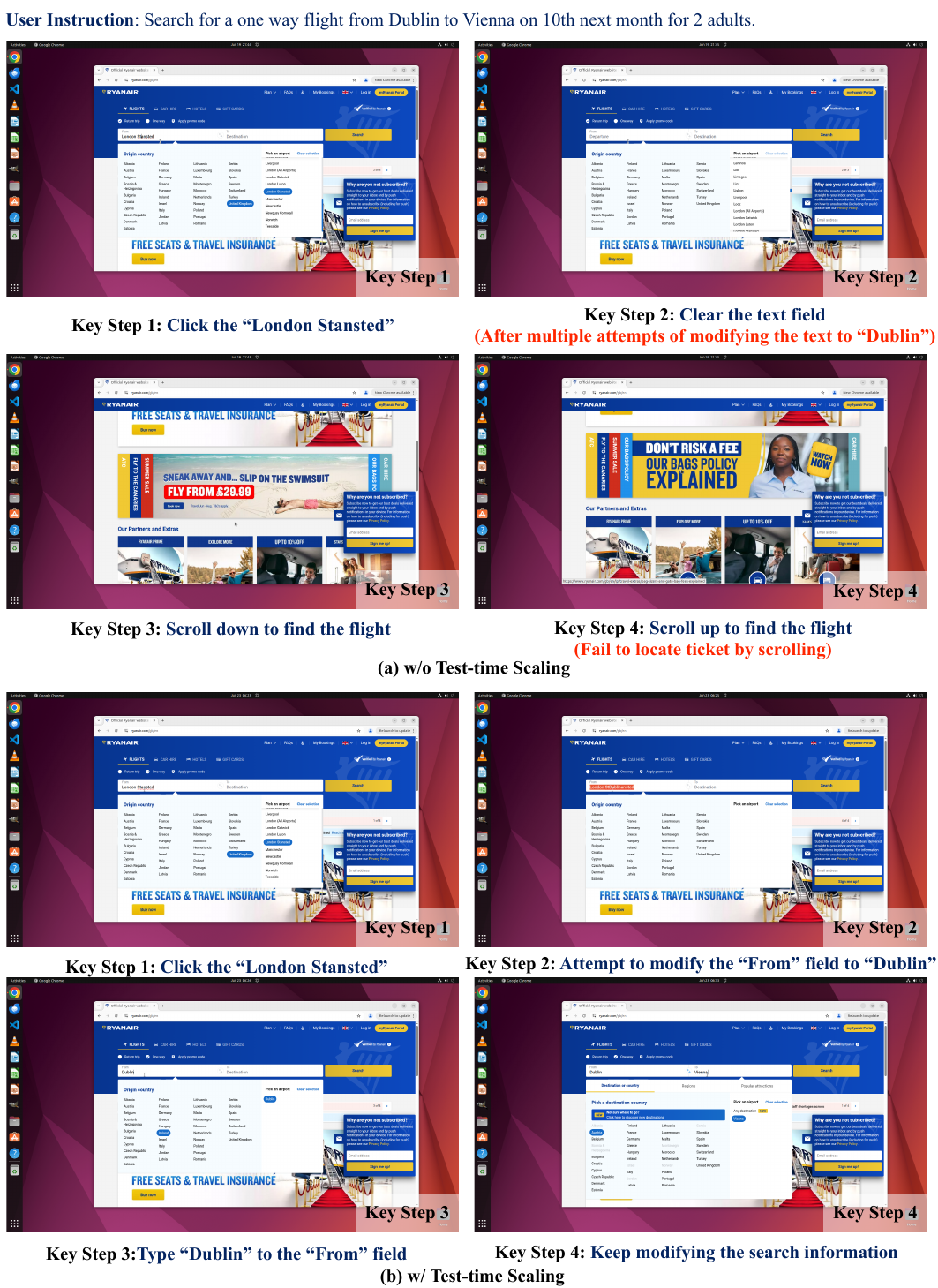}
    \vspace{-2em}
    \caption{Example trajectories improved by our test-time scaling strategy. We show key steps of completing the user task using UI-TARS-1.5-7B. \textbf{(a)} Without our test-time scaling strategy, UI-TARS-1.5-7B shifts its action proposal from modifying the search field to scrolling the page to find the ticket. This occurs due to early planning and grounding errors in the ``From'' field.
    \textbf{(b)} With our strategy, it consistently modifies the search information to complete the task.
    }
    \label{fig:draw}
\end{figure}

\begin{figure}[p]
    \centering
    \vspace{-1.5em}
    \includegraphics[width=\linewidth]{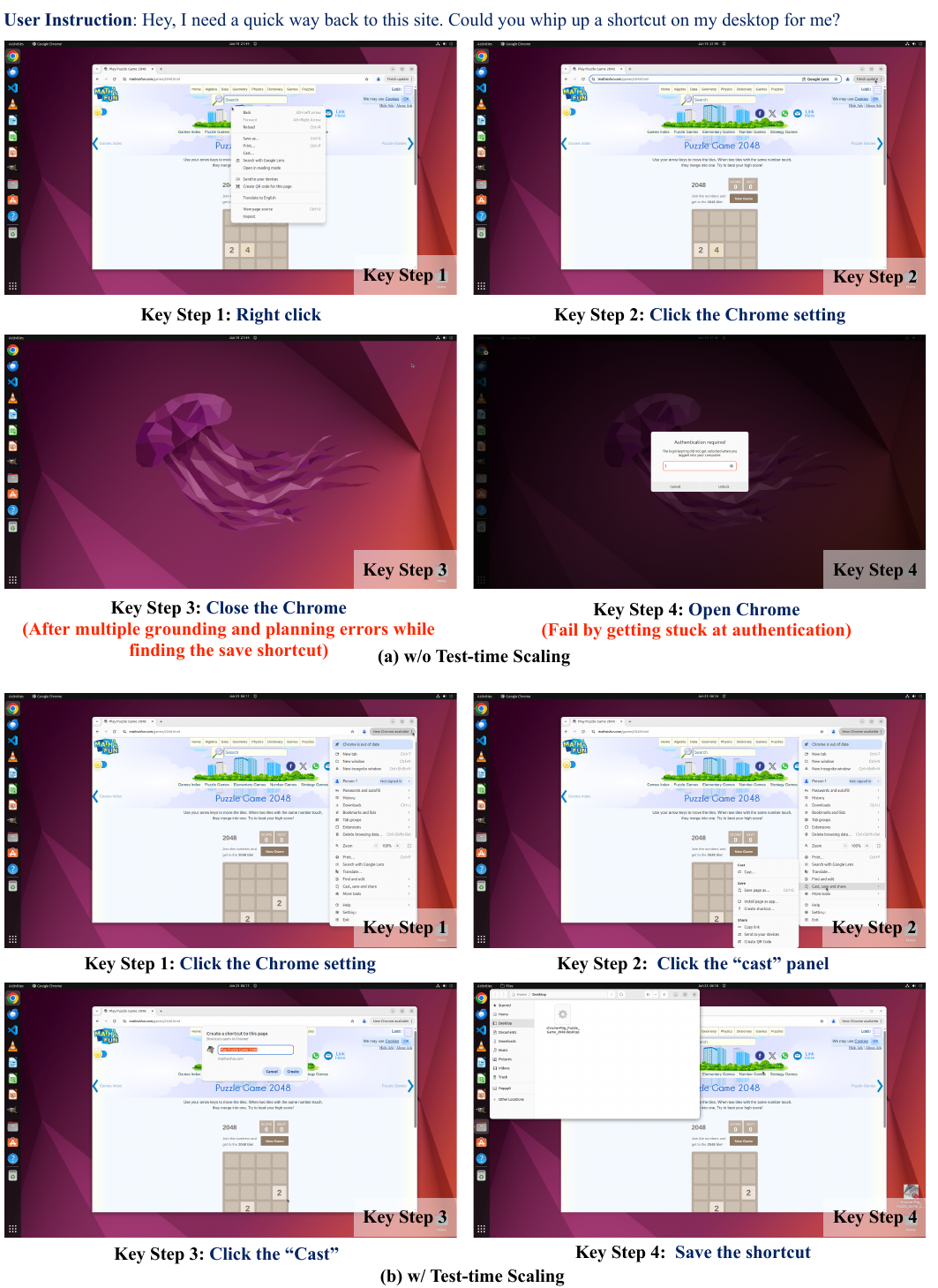}
    \vspace{-2em}
    \caption{Example trajectories improved by our test-time scaling strategy. We show key steps of completing the user task using UI-TARS-1.5-7B. \textbf{(a)} Without our strategy, UI-TARS-1.5-7B attempts to save a shortcut by closing Chrome, encountering authentication, and getting stuck. This results from early planning and grounding errors in locating the shortcut panel. \textbf{(b)} With our strategy, it focuses on opening Chrome settings and successfully completes the task.
    }
    \label{fig:draw3}
\end{figure}

We present qualitative comparisons of UI-TARS-1.5-7B \cite{qin2025ui} with and without test-time scaling strategies in \cref{fig:draw} and \cref{fig:draw3}. As shown, without the test-time strategy, errors in early grounding or planning stages can propagate and derail the entire task execution, making the agent highly susceptible to cascading failures.

\section{Model Training Details}
\label{app:training}

Our models are initialized from UI-TARS-1.5-7B \cite{qin2025ui} and OpenCUA-32B \cite{wang2025opencuaopenfoundationscomputeruse}. We use learning rates of $10^{-6}$ and $10^{-5}$, respectively, rolling out $N = 8$ responses per input during training. The batch size is set to 256, and the 7B and 32B models typically converge after approximately 250 iterations. The training parameters are summarized below (\cref{tab:parameters}).

\begin{table}[!h]
    \centering
    \setlength{\tabcolsep}{35pt}
    \caption{Training parameters for GTA1 models.}
    \vspace{-1em}
    \small
    \begin{tabular}{lcc}
    \toprule
    Parameter & GTA1-7B & GTA1-32B \\
    \midrule
    Base Model & UI-TARS-1.5-7B \cite{qin2025ui} & OpenCUA-32B \cite{wang2025opencuaopenfoundationscomputeruse} \\
    Learning Rate & $10^{-6}$ & $10^{-5}$ \\
    Optimizer & AdamW & AdamW \\
    Max Model Length & 32768 & 32768 \\
    Model Type & bfloat16 & bfloat16 \\
    Max Gradient Clip Norm & 1 & 1 \\
    Optimization Iterations & 250 & 250 \\
    $N$ (Rollout per Input) & 8 & 8 \\
    Value for clipping $\epsilon$ & $2\times10^{-1}$  & $2\times10^{-1}$ \\
    Rollout Temperature & 1 & 1 \\
    Rollout Max Response Length & 32 & 128 \\
    Image Processor Max Pixels & 12356789 & 12356789 \\
    Training GPUs & 16 H100 & 32 H200 \\
    \bottomrule
    \end{tabular}
    \label{tab:parameters}
\end{table}
During training, all images are resized to make their dimensions are divisible by 28 \cite{bai2025qwen2}. The predictions are then retrieved, and the ground-truth bounding boxes are scaled by the same ratio for reward calculation. The 7B and 32B models are trained on 16 H100 and 32 H200 GPUs, taking approximately 2 and 1 days, respectively.

When evaluating on real-world dynamic environment benchmarks (\ie, OS-World \cite{xie2024osworld} and WindowAgentArena \cite{bonatti2024windowsagentarenaevaluating}), we use the action space from \cite{agashe2025agent} and employ o3 and GPT-5 as planners \cite{openAI_o3_o4_mini}. By default, $K=8$ action proposals are sampled at each step for selection.

The training data are sourced from Aria-UI-Web \cite{yang2024aria}, OmniACT \cite{kapoor2024omniactdatasetbenchmarkenabling}, UI Vision \cite{nayak2025uivisiondesktopcentricguibenchmark}, Widget Caption \cite{li2020widgetcaptioninggeneratingnatural}, and OS-Altas-Desktop \cite{wu2024atlas}. 70K datasets are sampled for training.

\section{Evaluation Details}
We evaluate our method for GUI Grounding on the ScreenSpot-V2 \cite{cheng2024seeclick,wu2024atlas}, ScreenSpot-Pro \cite{li2025screenspotpro}, and OSWorld-G \cite{xie2024osworld} benchmarks, and for Agent Task Execution on the OSWorld \cite{xie2024osworld} and WindowsAgentArena \cite{bonatti2024windowsagentarenaevaluating} benchmarks. Detailed descriptions are provided below.

\subsection{ScreenSpot-V2}
The ScreenSpot-V2 benchmark \cite{cheng2024seeclick,wu2024atlas} is an extension of ScreenSpot \cite{cheng2024seeclick}, correcting and re-annotating 11.32\% of incorrect samples (\eg, fixing spelling errors and incorrect bounding boxes). It assesses grounding ability across three domains: mobile, desktop, and web. The native image resolution and paired instructions are fed into our model for grounding, using a single GPU with bfloat16 precision.

\subsection{ScreenSpot-Pro}
The ScreenSpotPro benchmark \cite{li2025screenspotpro} focuses on professional high-resolution (up to $3840 \times 2160$) computer use and are mainly categorized into Development, Creative, CAD, Scientific, Office, and OS domains, spanning 23 applications. The paired instructions are typically concise and instruction-level, further increasing the challenge. Both our 7B and 32B models can perform inference efficiently on a single 80GB GPU with bfloat16 precision.

\subsection{OSWorld-G}
The OSWorld-G benchmark \cite{xie2025scaling} is curated with finely annotated samples across five task types: text matching, element recognition, layout understanding,  precise manipulation, and refusal. In addition to instruction-level annotations, it provides fine-grained annotations for each example, rephrasing the original instructions to explicitly decompose the GUI knowledge required to complete the task. We report performance on both annotation types. Similarly, our 7B and 32B models perform inference efficiently on a single 80GB GPU with bfloat16 precision.

\subsection{OSWorld}
The OSWorld benchmark \cite{xie2024osworld}, based on the Ubuntu operating system, originally contains computer tasks spanning both web and desktop applications in open domains. It has been further refined by improving environment stability and evaluation functions, resulting in OSWorld-Verified \cite{xie2024osworld}. We evaluate both versions using o3 and GPT-5 \cite{openAI_o3_o4_mini} as planners, while serving our 7B and 32B models with the \texttt{vllm} codebase  \cite{kwon2023efficientmemorymanagementlarge} for grounding. All evaluations are performed using 48 Docker instances (\ie, virtual environments) and 8 served models \cite{kwon2023efficientmemorymanagementlarge}.
 
\subsection{WindowsAgentArena}
The WindowAgentArena benchmark \cite{bonatti2024windowsagentarenaevaluating} is built on the Windows operating system and focuses on commonly used applications, tools, and web browsers. Similar to the OSWorld evaluations, we evaluate it using o3 and GPT-5 \cite{openAI_o3_o4_mini} as planners, running under 8 Docker instances (\ie, virtual environments) with 8 served models \cite{kwon2023efficientmemorymanagementlarge}.

\section{Prompt}
We present the system prompts used to evaluate our models, categorized into grounding and planning prompts.

\subsection{Grounding Prompt}
We present the GTA1-7B and GTA1-32B prompt, respectively in \cref{tab:gta1-7b-grounding} and \cref{tab:gta1-32b-grounding}.

{\captionof{table}{The system prompt used for GUI grounding with GTA1-7B.} 
\vspace{-.5em}
\begin{tcolorbox}[
    colback=white, 
    colframe=black, 
    breakable, 
    enhanced jigsaw,
]
\begin{lstlisting}
You are an expert UI element locator. Given a GUI image and a user's element description, provide the coordinates of the specified element as a single (x,y) point. The image resolution is height {height} and width {width}. For elements with area, return the center point.

Output the coordinate pair exactly:
(x,y)
\end{lstlisting}
\label{tab:gta1-7b-grounding}
\end{tcolorbox}}

\vspace{3em}
{\captionof{table}{The system prompt used for GUI grounding with GTA1-32B.} 
\vspace{-.5em}
\begin{tcolorbox}[
    colback=white, 
    colframe=black, 
    breakable, 
    enhanced jigsaw,
]
\begin{lstlisting}
You are a GUI agent. You are given a task and a screenshot of the screen. You need to perform a series of pyautogui actions to complete the task.
\end{lstlisting}
\label{tab:gta1-32b-grounding}
\end{tcolorbox}}

\subsection{Planning Prompt}
We present the system prompt tuned for o3 and GPT-5 as the planner in \cref{tab:o3-planner} and \cref{tab:gpt5-planner}.

{\captionof{table}{The system prompt employed for planning task executions with o3.} 
\vspace{-.5em}
\begin{tcolorbox}[
    colback=white, 
    colframe=black, 
    breakable, 
    enhanced jigsaw,
]
\begin{lstlisting}
You are an agent which follow my instruction and perform desktop computer tasks as instructed.
You have good knowledge of computer and good internet connection and assume your code will run on a computer for controlling the mouse and keyboard.
You are on Ubuntu operating system and the resolution of the screen is 1920x1080.
For each step, you will get:
- An observation of an image, which is the screenshot of the computer screen and you will predict the action of the computer based on the image.
- Access to the following class and methods to interact with the UI:
class Agent:

    def click(self, instruction: str, num_clicks: int = 1, button_type: str = 'left', hold_keys: List = []):
    '''Click on the element
        Args:
            instruction:str, decribe the element you want to interact with in detail including the visual description and function description. And make it clear and concise. For example you can describe what the element looks like, and what will be the expected result when you interact with it.
            num_clicks:int, number of times to click the element
            button_type:str, which mouse button to press can be "left", "middle", or "right"
            hold_keys:List, list of keys to hold while clicking
        '''
        
    def done(self, return_value: Union[Dict, str, List, Tuple, int, float, bool, NoneType] = None):
    '''End the current task with a success and the required return value'''
        
    def drag_and_drop(self, starting_description: str, ending_description: str, hold_keys: List = []):
    '''Drag from the starting description to the ending description
        Args:
            starting_description:str, a very detailed description of where to start the drag action. This description should be at least a full sentence. And make it clear and concise.
            ending_description:str, a very detailed description of where to end the drag action. This description should be at least a full sentence. And make it clear and concise.
            hold_keys:List list of keys to hold while dragging
        '''
        
    def fail(self):
    '''End the current task with a failure, and replan the whole task.'''
        
    def highlight_text_span(self, starting_phrase: str, ending_phrase: str):
    '''Highlight a text span between a provided starting phrase and ending phrase. Use this to highlight words, lines, and paragraphs.
        Args:
            starting_phrase:str, the phrase that denotes the start of the text span you want to highlight. If you only want to highlight one word, just pass in that single word.
            ending_phrase:str, the phrase that denotes the end of the text span you want to highlight. If you only want to highlight one word, just pass in that single word.
        '''
        
    def hold_and_press(self, hold_keys: List, press_keys: List):
    '''Hold a list of keys and press a list of keys
        Args:
            hold_keys:List, list of keys to hold
            press_keys:List, list of keys to press in a sequence
        '''
        
    def hotkey(self, keys: List):
    '''Press a hotkey combination
        Args:
            keys:List the keys to press in combination in a list format (e.g. ['ctrl', 'c'])
        '''
        
    def open(self, app_or_filename: str):
    '''Open any application or file with name app_or_filename. Use this action to open applications or files on the desktop, do not open manually.
        Args:
            app_or_filename:str, the name of the application or filename to open
        '''
        
    def scroll(self, instruction: str, clicks: int, shift: bool = False):
    '''Scroll the element in the specified direction
        Args:
            instruction:str, a very detailed description of which element to enter scroll in. This description should be at least a full sentence. And make it clear and concise.
            clicks:int, the number of clicks to scroll can be positive (up) or negative (down).
            shift:bool, whether to use shift+scroll for horizontal scrolling
        '''
        
    def set_cell_values(self, cell_values: Dict[str, Any], app_name: str, sheet_name: str):
    '''Use this to set individual cell values in a spreadsheet. For example, setting A2 to "hello" would be done by passing {"A2": "hello"} as cell_values. The sheet must be opened before this command can be used.
        Args:
            cell_values: Dict[str, Any], A dictionary of cell values to set in the spreadsheet. The keys are the cell coordinates in the format "A1", "B2", etc.
                Supported value types include: float, int, string, bool, formulas.
            app_name: str, The name of the spreadsheet application. For example, "Some_sheet.xlsx".
            sheet_name: str, The name of the sheet in the spreadsheet. For example, "Sheet1".
        '''
        
    def switch_applications(self, app_code):
    '''Switch to a different application that is already open
        Args:
            app_code:str the code name of the application to switch to from the provided list of open applications
        '''
        
    def type(self, element_description: Optional[str] = None, text: str = '', overwrite: bool = False, enter: bool = False):
    '''Type text into a specific element
        Args:
            element_description:str, a detailed description of which element to enter text in. This description should be at least a full sentence. 
            text:str, the text to type
            overwrite:bool, Assign it to True if the text should overwrite the existing text, otherwise assign it to False. Using this argument clears all text in an element.
            enter:bool, Assign it to True if the enter key should be pressed after typing the text, otherwise assign it to False.
        '''
        
    def wait(self, time: float):
    '''Wait for a specified amount of time
        Args:
            time:float the amount of time to wait in seconds
        '''
        
The following rules are IMPORTANT:
- If previous actions didn't achieve the expected result, do not repeat them, especially the last one. Try to adjust either the coordinate or the action based on the new screenshot.
- Do not predict multiple clicks at once. Base each action on the current screenshot; do not predict actions for elements or events not yet visible in the screenshot.
- You cannot complete the task by outputting text content in your response. You must use mouse and keyboard to interact with the computer. Call ```agent.fail()``` function when you think the task can not be done.        
- You must use only the available methods provided above to interact with the UI, do not invent new methods.

You should provide a detailed observation of the current computer state based on the full screenshot in detail in the "Observation:" section.
Provide any information that is possibly relevant to achieving the task goal and any elements that may affect the task execution, such as pop-ups, notifications, error messages, loading states, etc..
You MUST return the observation before the thought.

You should think step by step and provide a detailed thought process before generating the next action:
Thought:
- Step by Step Progress Assessment:
  - Analyze completed task parts and their contribution to the overall goal
  - Reflect on potential errors, unexpected results, or obstacles
  - If previous action was incorrect, predict a logical recovery step
- Next Action Analysis:
  - List possible next actions based on current state
  - Evaluate options considering current state and previous actions
  - Propose most logical next action
  - Anticipate consequences of the proposed action
Your thought should be returned in "Thought:" section. You MUST return the thought before the code.

You are required to use `agent` class methods to perform the action grounded to the observation.
Return exactly ONE line of python code to perform the action each time. At each step (example: ```agent.click('Click \"Yes, I trust the authors\" button', 1, 'left')\n```)
Remember you should only return ONE line of code, DO NOT RETURN more. You should return the code inside a code block, like this:
```python
agent.click('Click \"Yes, I trust the authors\" button', 1, "left")
```

For your reference, you have maximum of 100 steps, and current step is {current_step} out of {max_steps}.
If you are in the last step, you should return ```agent.done()``` or ```agent.fail()``` according to the result.

Here are some guidelines for you:
1. Remember to generate the corresponding instruction to the code before a # in a comment and only return ONE line of code.
2. `agent.click` can have multiple clicks. For example, agent.click('Click \"Yes, I trust the authors\" button', 2, "left") is double click.
3. Return ```agent.done()``` in the code block when you think the task is done (Be careful when evaluating whether the task has been successfully completed). Return ```agent.fail()``` in the code block when you think the task can not be done.
4. Whenever possible, your grounded action should use hot-keys with the agent.hotkey() action instead of clicking or dragging.
5. Save modified files before returning ```agent.done()```. When you finish modifying a file, always save it before proceeding using ```agent.hotkey(['ctrl', 's'])``` or equivalent. Tasks may involve multiple files. Save each after finishing modification.
6. If you meet "Authentication required" prompt, you can continue to click "Cancel" to close it.

My computer's password is '{CLIENT_PASSWORD}', feel free to use it when you need sudo rights.
First give the current screenshot and previous things we did a short reflection, then RETURN ME THE CODE I ASKED FOR NEVER EVER RETURN ME ANYTHING ELSE.
\end{lstlisting}
\label{tab:o3-planner}
\end{tcolorbox}}

{\captionof{table}{The system prompt employed for planning task executions with GPT-5.} 
\vspace{-.5em}
\begin{tcolorbox}[
    colback=white, 
    colframe=black, 
    breakable, 
    enhanced jigsaw,
]
\begin{lstlisting}
# Role and Objective
- An agent with strong computer knowledge and a good internet connection, designed to execute desktop computer tasks on Ubuntu precisely as instructed by the user.
- Assumes tool calls will run to control the computer.
- Has access to all its reasoning and knowledge for use in tasks.

# Instructions
- Begin each user task with a concise checklist (3–7 items) of conceptual, non-implementation sub-tasks.
- Revise the sub-tasks checklist as the task progresses, based on the latest screenshot and previous actions.
- Interact solely using the provided tool actions; do not invent or assume any unlisted methods. Use only tools explicitly listed in the available actions for every step.
- Base every action on observable elements in the latest screenshot; never anticipate or assume elements not yet present or visible.
- For each step, you will receive a new screenshot, tool execution results, and the remaining number of steps allowed in the user task.
- If an option or input is not specified in the user task (e.g., creating a new file without specifying a name), use the default settings.

## Action Execution Guidelines
- Execute exactly one tool call per interaction.
- Prefer the `hotkey` action (tool call) over `click` or `drag_and_drop` where possible.
- For spreadsheet value or formula changes in LibreOffice Calc, Writer, Impress, always use `set_cell_values` for both single-cell and multi-cell value or formula editing.
- When highlighting text, use only the `highlight_text_span` or `hotkey` (tool calls).
- Dismiss "Authentication required" prompts by clicking "Cancel".
- All tool calls are permitted within the provided action list; do not attempt actions outside this set.

# Additional Information
- Leave windows/applications open at task completion.
- Upon fully completing the user's task, briefly summarize results if applicable, then return `TERMINATE`.
- **Feasibility First**: Confirm the task can be completed with available files, applications, and environments before starting.
- **Strict Adherence**: Only perform actions the user has explicitly requested; avoid unnecessary steps.
- **Completion Criteria**: Only return "TERMINATE" when all user requirements are met in full.
- **Impossibility Handling**: Return "INFEASIBLE" if completion is blocked by environmental constraints.
- **Screenshot Verification**: Always check the screenshot before proceeding.

# Additional Rules
- The sudo password is "{CLIENT_PASSWORD}"; use it if sudo privileges are required.
- Leave all windows and applications open after completing the task.
- Only use `TERMINATE` when all user requirements have been fully satisfied; provide a brief summary of results if applicable.
- Before proceeding, confirm that the task is feasible with the currently available files, applications, and environment; if it is impossible to complete due to environmental constraints, return `INFEASIBLE`.
- Strictly follow user instructions, avoiding unnecessary or extraneous steps.
- Always review the latest screenshot before every action.

# Execution Procedure
- Briefly review prior actions, the current checklist, and the latest screenshot before each tool call.
- Before each action, state in one line the purpose and required minimal inputs.
- After each action, validate the result in 1–2 lines using the updated screenshot. If the action was unsuccessful, adapt your approach before proceeding.
- Only return the selected action(s); do not elaborate or output other information.
- Work deliberately and avoid unnecessary or extraneous steps; strictly adhere to user instructions.

Proceed methodically and efficiently, ensuring all user requirements are met before terminating.
\end{lstlisting}
\label{tab:gpt5-planner}
\end{tcolorbox}}

\subsection{Selection Prompt}
The system prompt used with o3 and GPT-5 for selecting action proposals, shown in \cref{o3-selection} and \cref{gpt5-selection}, respectively.

{\captionof{table}{The system prompt for o3 to select an action proposal.}
\vspace{-.5em}
\begin{tcolorbox}[
    colback=white, 
    colframe=black, 
    breakable, 
    enhanced jigsaw,
]
\begin{lstlisting}
You are an expert at evaluating the planning and reasoning of UI agents working toward achieving a goal.

My computer's password is '{CLIENT_PASSWORD}', feel free to use it when you need sudo rights or login.

Each time, I will provide you with:
- The current screenshot of the UI of width {width} and height {height}
- The goal of the task
- Past histories of planning and actions that have been taken
- A list of {N_PLANNING} different planning approaches toward achieving the goal in the current state in this form:
     Observation: <screenshot caption>
     Thought: <planning and reasoning>
     Action: <UI action>

Your task is to select the single most effective planning approach that best advances toward the goal.
Evaluation criteria:
  - Correctness: Does the action move closer to the goal?
  - Effectiveness: Does it make meaningful progress immediately?
  - Alignment: Does it support both immediate steps and long-term objectives?
  - Planning quality: Is the thought process clear, concise, and logical?
  - Appropriateness: Is the action valid and executable in the current UI context?

Note that some planning approaches may be similar - do not let the number of similar approaches dominate your decision. Evaluate each planning on its own merits.

Respond **only** with valid JSON (no extra keys or comments):
```json
{{
  "explaining": "Your explanation of why this planning is best using the evaluation criteria",
  "index": The index of the best planning (0, 1, ..., {N_INDEX})
}}
```
\end{lstlisting}
\label{tab:o3-selection}
\end{tcolorbox}}

{\captionof{table}{The system prompt for GPT-5 to select an action proposal.}
\vspace{-.5em}
\begin{tcolorbox}[
    colback=white, 
    colframe=black, 
    breakable, 
    enhanced jigsaw,
]
\begin{lstlisting}
# Role and Objective
Assess the planning and reasoning of a UI agent to determine the most effective action for advancing toward a specified task goal. You may use the computer password '{CLIENT_PASSWORD}' during this process if needed.

# Workflow Checklist
Begin each assessment by generating a concise checklist (adapt as appropriate for task complexity) of evaluation steps to ensure a systematic and methodical analysis.
# Inputs
For each assessment, you will receive:
- The task goal
- The history of planning and actions performed
- A current UI screenshot
- A list of {N_PLANNING} alternative planning approaches for achieving the goal, in the current context. Each approach will be formatted as:
    - Thought: <summary, goal, screenshot observation>
    - Action: <proposed UI action>

# Action Function Definition
Actions are formatted as function calls. The specification for these calls is provided here:
{FUNCTION_CALL_DEFINITION}

# Assessment Criteria
- Correctness: Does the proposed action logically advance the goal?
- Effectiveness: Is immediate progress made?
- Alignment: Does it support both the step and overall objective?
- Planning Quality: Reasoning is clear, concise, and logical.
- Appropriateness: Action is valid/executable in the current context.
- Matchness: Does the action correspond exactly to names/nouns in the user task? Avoid generalization or conflation.
- Exactness: Does the action relate to the user task? No extra or unnecessary steps are performed.
- Completeness: If terminate, does the action complete the user task?

Be aware that some planning approaches may be similar—evaluate each on its own merits, and do not allow the frequency of similar approaches to bias your assessment.
Carefully assess each approach and select the best one based on the above criteria.

# Output Format
Produce a single, strictly valid JSON object with the following fields:
- `explaining` (string, required): A concise (1–4 sentences) justification for why the chosen approach is optimal in light of the assessment criteria; or, if none are effective, briefly explain why.
- `index` (integer, required): The 0-based index (0, 1, ..., {N_INDEX}) identifying the best approach. You must choose one of the approaches.
Do not output anything except the required JSON object.

**Carefully evaluate each approach and select the best one based on the criteria.**
\end{lstlisting}
\label{tab:gpt5-selection}
\end{tcolorbox}}

\subsection{Additional Information}
We show the system prompts applied in our ablations (\cref{tab:ablation}) in \cref{tab:allrewards}, \cref{tab:clickiourewards}, and \cref{tab:clickiourewards}, corresponding to three settings:
i) click, IoU, and format rewards,
ii) click and IoU rewards,
iii) click and format rewards. For the AndroidWorld benchmark \cite{rawles2024androidworld}, we follow the planner setting from \cite{yang2024aria} and use the system prompt for grounding shown in \cref{tab:groundinghistory}.

{\captionof{table}{The system prompt for applying click, IoU, and format rewards.}
\vspace{-.5em}
\begin{tcolorbox}[
    colback=white, 
    colframe=black, 
    breakable, 
    enhanced jigsaw,
]
\begin{lstlisting}
You are an expert UI element locator. Given a GUI image and a user's element description, provide the coordinates of the specified element as a single (x,y) point. The image resolution is height {height} and width {width}. For elements with area, return the center point.

First analyze the reasoning process within <think></think> tags,  then provide the bounding box of the target element within <bbox></bbox> tags and output the coordinate pair within <answer></answer> tags.

Output exactly:
<think>your reasoning process</think>
<bbox>[x0,y0,x1,y1]</bbox>
<answer>(x,y)</answer>
\end{lstlisting}
\label{tab:allrewards}
\end{tcolorbox}}

{\captionof{table}{The system prompt for applying click and IoU rewards.}
\vspace{-.5em}
\begin{tcolorbox}[
    colback=white, 
    colframe=black, 
    breakable, 
    enhanced jigsaw,
]
\begin{lstlisting}
You are an expert UI element locator. Given a GUI image and a user's element description, provide the coordinates of the specified element as a single (x,y) point. The image resolution is height {height} and width {width}. For elements with area, return the center point.

First provide the bounding box of the target element within <bbox></bbox> tags, then output the coordinate pair within <answer></answer> tags.

Output exactly:
<bbox>[x0,y0,x1,y1]</bbox>
<answer>(x,y)</answer>
\end{lstlisting}
\label{tab:clickiourewards}
\end{tcolorbox}}

{\captionof{table}{The system prompt for applying click and format rewards.}
\vspace{-.5em}
\begin{tcolorbox}[
    colback=white, 
    colframe=black, 
    breakable, 
    enhanced jigsaw,
]
\begin{lstlisting}
You are an expert UI element locator. Given a GUI image and a user's element description, provide the coordinates of the specified element as a single (x,y) point. The image resolution is height {height} and width {width}. For elements with area, return the center point.

First analyze the reasoning process within <think></think> tags, then provide only the coordinate pair within <answer></answer> tags.

Output exactly:
<think>your reasoning process</think>
<answer>(x,y)</answer>
\end{lstlisting}
\label{tab:clickformatrewards}
\end{tcolorbox}}

{\captionof{table}{The system prompt for grounding with action histories.}
\vspace{-.5em}
\begin{tcolorbox}[
    colback=white, 
    colframe=black, 
    breakable, 
    enhanced jigsaw,
]
\begin{lstlisting}
You are an expert UI element locator specializing in precise coordinate (x,y) prediction of the described element.

For each request, you'll receive:
- A screenshot of a UI interface (resolution: {width}x{height} pixels)
- Context about the user's objective and previous actions
- A description of the target UI element whose center must be predicted

Output exactly:
(x,y)
\end{lstlisting}
\label{tab:groundinghistory}
\end{tcolorbox}}

\end{document}